# Counting-Based Search: Branching Heuristics for Constraint Satisfaction Problems

**Gilles Pesant** GILLES.PESANT@POLYMTL.CA
*École polytechnique de Montréal, Montréal, Canada*

**Claude-Guy Quimper** CLAUDE-GUY.QUIMPER@IFT.ULAVAL.CA
*Université Laval, Québec, Canada*

**Alessandro Zanarini** ALESSANDRO.ZANARINI@DYNADEC.COM
*Dynadec Europe, Belgium*

## Abstract

Designing a search heuristic for constraint programming that is reliable across problem domains has been an important research topic in recent years. This paper concentrates on one family of candidates: counting-based search. Such heuristics seek to make branching decisions that preserve most of the solutions by determining what proportion of solutions to each individual constraint agree with that decision. Whereas most generic search heuristics in constraint programming rely on local information at the level of the individual variable, our search heuristics are based on more global information at the constraint level. We design several algorithms that are used to count the number of solutions to specific families of constraints and propose some search heuristics exploiting such information. The experimental part of the paper considers eight problem domains ranging from well-established benchmark puzzles to rostering and sport scheduling. An initial empirical analysis identifies heuristic maxSD as a robust candidate among our proposals. We then evaluate the latter against the state of the art, including the latest generic search heuristics, restarts, and discrepancy-based tree traversals. Experimental results show that counting-based search generally outperforms other generic heuristics.

## 1. Introduction

Constraint Programming (CP) is a powerful technique to solve combinatorial problems. It applies sophisticated inference to reduce the search space and a combination of variable- and value-selection heuristics to guide the exploration of that search space. Because this inference is encapsulated in each constraint appearing in the model of a problem, users may consider it as a black box. In contrast, search in CP is programmable, which is a mixed blessing. It allows one to easily tailor search to a problem, adding expertise and domain knowledge, but it may also discourage the average user who would prefer a generic and fairly robust default search heuristic that works well most of the time. Some generic search heuristics are indeed available in CP but robustness remains an issue.

Whereas most generic search heuristics in constraint programming rely on information at the level of the individual variable (e.g. its domain size and degree in the constraint network), we investigate search heuristics based on more global information. "Global" constraints in CP are successful because they encapsulate powerful dedicated inference algorithms but foremost because they bring out the underlying structure of combinatorial problems. That exposed structure can also be exploited during search. Search heuristics following the *fail-first principle* (detect failure as early as possible) and centered on constraints can be guided by a count of the number of solutions left for each constraint. We might for example focus search on the constraint currently having the smallest number of solutions, recognizing that failure necessarily occurs through a constraint admitting no more solution. We can also count the number of solutions featuring a given variable-value assignment in an individual constraint, favoring assignments appearing in a high proportion of solutions with the hope that such a choice generally brings us closer to satisfying the whole CSP.

The concept of counting-based search heuristics has already been introduced, most recently by Zanarini and Pesant (2009). The specific contributions of this paper are: additional counting algorithms, including for other families of constraints, thus broadening the applicability of these heuristics; experiments that include the effect of some common features of search heuristics such as search tree traversal order, restarts and learning; considerable empirical evidence that counting-based search outperforms other generic heuristics.

In the rest of the paper: Section 2 provides background and reviews related work; Sections 3 to 5 present counting algorithms for several of the most usual constraints; Section 6 introduces counting-based search heuristics which can exploit the algorithms of the previous sections; Section 7 reports on an extensive experimental study comparing our proposed heuristics to state-of-the-art generic heuristics on many problem domains; finally Section 8 concludes the paper.

## 2. Background and Related Work

We start with the usual general representation formalism for CP.

**Definition 1** (constraint satisfaction problem (CSP))**.** *Given a finite set of variables* $X = \{x_1, x_2, \ldots\}$*, a finite domain of possible values for each of these variables,* $\mathcal{D} = \{D_1, \ldots, D_{|X|}\}$*,* $x_i \in D_i \quad (1 \leq i \leq |X|)$*, and a finite set of constraints (relations) over subsets of* $X$*,* $C = \{c_1, c_2, \ldots\}$*, the* constraint satisfaction problem $(X, \mathcal{D}, C)$ *asks for an assignment of a value from* $D_i$ *to each variable* $x_i$ *of* $X$ *that satisfies (belongs to) each* $c_j$ *in* $C$*.*

And now recall some definitions and notation from Pesant (2005) and Zanarini and Pesant (2009).

**Definition 2** (solution count)**.** *Given a constraint* $c(x_1, \ldots, x_n)$ *and respective finite domains* $D_i \;_{1 \leq i \leq n}$*, let* $\#c(x_1, \ldots, x_n)$ *denote the number of* $n$*-tuples in the corresponding relation, called its* solution count*.*

**Definition 3** (solution density)**.** *Given a constraint* $c(x_1, \ldots, x_n)$*, respective finite domains* $D_i \;_{1 \leq i \leq n}$*, a variable* $x_i$ *in the scope of* $c$*, and a value* $d \in D_i$*, we will call*

$$\sigma(x_i, d, c) = \frac{\#c(x_1, \ldots, x_{i-1}, d, x_{i+1}, \ldots, x_n)}{\#c(x_1, \ldots, x_n)}$$

*the* solution density *of pair* $(x_i, d)$ *in* $c$*. It measures how often a certain assignment is part of a solution to* $c$*.*

Heuristics are usually classified in two main categories: static variable ordering heuristics (SVOs) and dynamic variable ordering heuristics (DVOs). The former order the variables prior to search and do not revise the ordering during search. Common SVOs are lexicographic order, lexico, and decreasing degree (i.e. number of constraints in which a variable is involved), deg. DVOs are generally considered more effective as they exploit information gathered during search. They often follow the *fail-first principle* originally introduced by Haralick and Elliott (1980, p. 263) i.e. "To succeed, try first where you are most likely to fail." The same authors proposed the widely-used heuristic dom that branches on the variables with the smallest domain; the aim of such a heuristic is to minimize branch depth. A similar heuristic, proposed by Brélaz (1979), selects the variable with the smallest remaining domain and then breaks ties by choosing the one with the highest *dynamic degree* - ddeg [1] (that is, the one constraining the largest number of unbound variables). Bessière and Régin (1996) and Smith and Grant (1998) combined the domain and degree information by minimizing the ratio dom/deg or dom/ddeg.

### 2.1 Impact-Based Heuristics

Refalo (2004) proposed Impact Based Search (IBS), a heuristic that chooses the variable whose instantiation triggers the largest search space reduction (highest impact) that is approximated as the reduction of the product of the variable domain cardinalities. More formally the impact of a variable-value pair is:

$$I(x_i = d) = 1 - \frac{P_{after}}{P_{before}}$$

where $P_{after}$ and $P_{before}$ are the products of the domain cardinalities respectively after and before branching on $x_i = d$ (and propagating that decision). The impact is either computed exactly at a given node of the search (the exact computation provides better information but is more time consuming) or approximated as the average reduction observed during the search (hence automatically collected on-the-go at almost no additional cost), that is:

$$\bar{I}(x_i = d) = \frac{\sum_{k \in K} I^k(x_i = d)}{|K|}$$

where $K$ is the index set of the impact observed so far for the assignment $x_i = d$. The variable impact is defined by Refalo (2004) as

$$\mathcal{I}(x_i) = \sum_{d \in D'_i} 1 - \bar{I}(x_i = d)$$

where $D'_i$ is the current domain of the variable $x_i$. Impact initialization is fundamental to obtain a good performance even at the root of the search tree; therefore, Refalo proposed to initialize the impacts by probing each variable-value pair at the root node (note that this subsumes a reduced form of singleton consistency at the root node and can be quite computationally costly). IBS selects the variable having the largest impact (hence trying

1. It is also referred to as *future degree* or *forward degree* in the literature.

to maximize the propagation effects and the reduction of the search space) and then selects the value having the smallest impact (hence leaving more choices for the future variables).

As an interesting connection with impact-based heuristics, Szymanek and O'Sullivan (2006) proposed to query the model constraints to approximate the number of filtered values by each constraint individually. This information is then exploited to design a variable and/or value selection heuristic. Nonetheless, it differs from impact-based search as they take into consideration each constraint separately, and from counting-based heuristics (Zanarini & Pesant, 2009) as the information provided is more coarse-grained than actual solution counts.

### 2.2 Conflict-Driven Heuristics

Boussemart, Hemery, Lecoutre, and Sais (2004) proposed a conflict-driven variable ordering heuristic: they extended the concept of *variable degree* integrating a simple but effective learning technique that takes failures into account. Basically each constraint has an associated weight that is increased by one each time the constraint leads to a failure (i.e. a domain wipe-out). A variable has a *weighted degree* – wdeg – that is the sum of the weights of constraints in which it is involved. Formally, the weighted degree of a variable is:

$$\alpha_{wdeg}(x_i) = \sum_{c \in C} weight[c] \quad | \quad Vars(c) \ni x_i \wedge |FutVars(c)| > 1$$

where $FutVars(c)$ denotes the uninstantiated variables of the constraint $c$, $weight[c]$ is its weight and $Vars(c)$ the variables involved in $c$. The heuristics proposed simply choose the variable that maximizes wdeg or minimizes dom/wdeg. These heuristics offer no general method to deal with global constraints: a natural extension is to increase the weight of every variable in a failed constraint but most of them may not have anything to do with the failure, which dilutes the conflict information. They are also particularly sensitive to revision orderings (i.e. the ordering of the propagation queue) hence leading to varying performance. Grimes and Wallace (2006, 2007) proposed some adaptations of dom/wdeg when combined with restarts or by updating weights on value deletions as well. Balafoutis and Stergiou (2008b) proposed, among other improvements over the original dom/wdeg, *weight aging*, that is the constraint weights are periodically reduced. This limits the inertia of constraints that got a significant weight early in the search but that are not critical anymore later on.

Nowadays heuristics dom/wdeg and IBS are considered to be the state of the art of generic heuristics with no clear dominance by one or the other (Balafoutis & Stergiou, 2008a). Finally we note that both rely on the hypothesis that what is learned early in the search will tend to remain true throughout the search tree: impacts should not change much from one search tree node to the other; the same constraints lead to domain wipe-outs in different parts of the search tree.

### 2.3 Approximated Counting-Based Heuristics

The idea of using an approximation on the number of solutions of a problem as heuristic is not new. Kask, Dechter, and Gogate (2004) approximate the total number of solutions extending a partial solution to a CSP and use it in a value selection heuristic, choosing the value whose assignment to the current variable gives the largest approximate solution count. An implementation optimized for binary constraints performs well compared to other popular strategies. Hsu, Kitching, Bacchus, and McIlraith (2007) and later Bras, Zanarini, and Pesant (2009) apply a Belief Propagation algorithm within an Expectation Maximization framework (EMBP) in order to approximate variable biases (or marginals) i.e. the probability a variable takes a given value in a solution. The resulting heuristics tend to be effective but quite time-consuming. One way to differentiate our work from these is that we focus on fine-grained information from individual constraints whereas they work on coarser information over the whole problem.

## 3. Counting for Alldifferent Constraints

The `alldifferent` constraint restricts a set of variables to be pairwise different (Régin, 1994).

**Definition 4** (Alldifferent Constraint). *Given a set of variables $X = \{x_1, \ldots, x_n\}$ with respective domains $D_1, \ldots, D_n$, the set of tuples allowed by* `alldifferent`$(X)$ *are:*

$$\{(d_1, d_2, \ldots, d_n) \mid d_i \in D_i, d_i \neq d_j \forall i \neq j\}$$

We define the associated (0-1) square matrix $A = (a_{id})$ with $|\bigcup_{i=1,\ldots,n} D_i|$ rows and columns such that $a_{id} = 1$ iff $d \in D_i$ [2]. If there are more distinct values in the domains than there are variables, say $p$ more, we add $p$ rows filled with 1s to matrix $A$. An equivalent representation is given by the bipartite *value graph* with a vertex for each variable and value and edges corresponding to "1" entries in $A$.

Then as discussed by Zanarini and Pesant (2009), counting the number of solutions to an `alldifferent` constraint is equivalent to computing the *permanent* of $A$ (or the number of maximum matchings in the value graph), formally defined as

$$perm(A) = \sum_{d=1}^{n} a_{1,d}\, perm(A_{1,d}) \tag{1}$$

where $A_{1,d}$ denotes the submatrix obtained from $A$ by removing row 1 and column $d$ (the permanent of the empty matrix is equal to 1). If $p$ extra rows were added, the result must be divided by $p!$ as shown by Zanarini and Pesant (2010).

Because computing the permanent is well-known to be $\#P$-complete (Valiant, 1979), Zanarini and Pesant (2009) developed an approach based on sampling which gave close approximations and led to very effective heuristics on hard instances. However it was not competitive on easy to medium difficulty instances because of the additional computational effort. The next section describes an approach based on upper bounds, trading approximation accuracy for a significant speedup in the counting procedure.[3]

2. For notational convenience and without loss of generality, we identify domain values with consecutive natural numbers.
3. This was originally introduced by Zanarini and Pesant (2010).

### 3.1 Upper Bounds

In the following we assume for notational convenience that matrix $A$ has $n$ rows and columns and we denote by $r_i$ the sum of the elements in the $i^{th}$ row of $A$ (i.e. $r_i = \sum_{d=1}^{n} a_{id}$).

A first upper bound for the permanent was conjectured by Minc (1963) and later proved by Brégman (1973):

$$perm(A) \leq \prod_{i=1}^{n} (r_i!)^{1/r_i}. \tag{2}$$

Recently Liang and Bai (2004) proposed a second upper bound (with $q_i = min\{\lceil \frac{r_i+1}{2} \rceil, \lceil \frac{i}{2} \rceil\}$):

$$perm(A)^2 \leq \prod_{i=1}^{n} q_i(r_i - q_i + 1). \tag{3}$$

Neither of these two upper bounds strictly dominates the other. In the following we denote by $UB^{BM}(A)$ the Brégman-Minc upper bound and by $UB^{LB}(A)$ the Liang-Bai upper bound. Jurkat and Ryser (1966) proposed another bound:

$$perm(A) \leq \prod_{i=1}^{n} min(r_i, i).$$

However it is considered generally weaker than $UB^{BM}(A)$ (see Soules, 2005 for a comprehensive literature review).

#### 3.1.1 Algorithm

We decided to adapt $UB^{BM}$ and $UB^{LB}$ in order to compute an approximation of solution densities for the `alldifferent` constraint. Assigning $d$ to variable $x_i$ translates to replacing the $i^{th}$ row by the unit vector $e(d)$ (i.e. setting the $i^{th}$ row of the matrix to 0 except for the element in column $d$). We write $A_{x_i=d}$ to denote matrix $A$ except that $x_i$ is fixed to $d$. We call *local probe* the assignment $x_i = d$ performed to compute $A_{x_i=d}$ i.e. a temporary assignment that does not propagate to any other constraint except the one being processed.

The upper bound on the number of solutions of the `alldifferent`$(x_1, \ldots, x_n)$ constraint with a related adjacency matrix $A$ is then simply

$$\#\texttt{alldifferent}(x_1, \ldots, x_n) \leq min\{UB^{BM}(A), UB^{LB}(A)\}$$

Note that in Formulas 2 and 3, the $r_i$'s are equal to $|D_i|$; since $|D_i|$ ranges from 0 to $n$, the factors can be precomputed and stored: in a vector $BMfactors[r] = (r!)^{1/r}, r = 0, \ldots, n$ for the first bound and similarly for the second one (with factors depending on both $|D_i|$ and $i$). Assuming that $|D_i|$ is returned in $O(1)$, computing the formulas takes $O(n)$ time. Solution densities are then approximated as

$$\sigma(x_i, d, \texttt{alldifferent}) \approx \frac{min\{UB^{BM}(A_{x_i=d}), UB^{LB}(A_{x_i=d})\}}{\eta}$$

where $\eta$ is a normalizing constant so that $\sum_{d \in D_i} \sigma(x_i, d, \texttt{alldifferent}) = 1$.

The local probe $x_i = d$ may trigger some local propagation according to the level of consistency we want to achieve; therefore $A_{x_i=d}$ is subject to the filtering performed on the constraint being processed. Since the two bounds in Formulas 2 and 3 depend on $|D_i|$, a stronger form of consistency would likely lead to more changes in the domains and on the bounds, and presumably to more accurate solution densities.

If we want to compute $\sigma(x_i, d, \texttt{alldifferent})$ for all $i = 1, \ldots, n$ and for all $d \in D_i$ then a trivial implementation would compute $A_{x_i=d}$ for each variable-value pair; the total time complexity would be $O(mP + mn)$ (where $m$ is the sum of the cardinalities of the variable domains and $P$ the time complexity of the filtering).

Although unable to improve over the worst case complexity, in the following we propose an algorithm that performs definitely better in practice. We introduce before some additional notation: we write as $D'_k$ the variable domains after enforcing $\theta$-consistency[4] on that constraint alone and as $\tilde{I}_{x_i=d}$ the set of indices of the variables that were subject to a domain change due to a local probe and the ensuing filtering, that is, $k \in \tilde{I}_{x_i=d}$ iff $|D'_k| \neq |D_k|$. We describe the algorithm for the Brégman-Minc bound — it can be easily adapted for the Liang-Bai bound.

The basic idea is to compute the bound for matrix $A$ and to reuse it to speed up the computation of the bounds for $A_{x_i=d}$ for all $i = 1, \ldots, n$ and $d \in D_i$. Let

$$\gamma_k = \begin{cases} \frac{BMfactors[|D'_k|]}{BMfactors[|D_k|]} & \text{if } k \in \tilde{I}_{x_i=d} \\ \\ 1 & \text{otherwise} \end{cases}$$

$$\begin{aligned} UB^{BM}(A_{x_i=d}) &= \prod_{k=1}^{n} BMfactors[|D'_k|] = \prod_{k=1}^{n} \gamma_k \; BMfactors[|D_k|] \\ &= UB^{BM}(A) \prod_{k=1}^{n} \gamma_k \end{aligned}$$

Note that $\gamma_k$ with $k = i$ (i.e. we are computing $UB^{BM}(A_{x_i=d})$) does not depend on $d$; however $\tilde{I}_{x_i=d}$ does depend on $d$ because of the domain filtering.

Algorithm 1 shows the pseudo code for computing $UB^{BM}(A_{x_i=d})$ for all $i = 1, \ldots, n$ and $d \in D_i$. Initially, it computes the bound for matrix $A$ (line 1); then, for a given $i$, it computes $\gamma_i$ and the upper bound is modified accordingly (line 3). Afterwards, for each $d \in D_i$, $\theta$-consistency is enforced (line 7) and it iterates over the set of modified variables (line 9-10) to compute all the $\gamma_k$ that are different from 1. We store the upper bound for variable $i$ and value $d$ in the structure $VarValUB[i][d]$. Before computing the bound for the other variables-values the assignment $x_i = d$ needs to be undone (line 12). Finally, we normalize the upper bounds in order to correctly return solution densities (line 13-14). Let $\mathcal{I}$ be equal to $\max_{i,d} |\tilde{I}_{x_i=d}|$, the time complexity is $O(mP + m\mathcal{I})$.

If matrix $A$ is dense we expect $\mathcal{I} \simeq n$. Therefore most of the $\gamma_k$ are different from 1 and need to be computed. As soon as the matrix becomes sparse enough then $\mathcal{I} \ll n$ and only a small fraction of $\gamma_k$ need to be computed, and that is where Algorithm 1 has an advantage.

4. Stands for any form of consistency

```
UB = UB^{BM}(A);
for i = 1, ..., n do
    varUB = UB * BMfactors[1] / BMfactors[|D_i|];
    total = 0;
    forall d ∈ D_i do
        set x_i = d;
        enforce θ-consistency;
        VarValUB[i][d] = varUB;
        forall k ∈ Ĩ_{x_i=d} \ {i} do
            VarValUB[i][d] = VarValUB[i][d] * BMfactors[|D'_k|] / BMfactors[|D_k|];
        total = total + VarValUB[i][d];
        rollback x_i = d;
    forall d ∈ D_i do
        SD[i][d] = VarValUB[i][d]/total;
return SD;
```

**Algorithm 1**: Solution Densities

The sampling algorithm introduced by Zanarini and Pesant (2009) performed very well both in approximating the solution count and the solution densities, but this is not the case for upper bounds. The latter in fact produce weak approximations of the solution count but offer a very good trade-off between performance and accuracy for solution densities: taking the ratio of two solution counts appears to cancel out the weakness of the original approximations (see Zanarini & Pesant, 2010 for further details).

### 3.2 Symmetric Alldifferent

Régin (1999) proposed the `symmetric_alldifferent` constraint that is a special case of the `alldifferent` in which variables and values are defined from the same set. This is equivalent to a traditional `alldifferent` with an additional set of constraints stating that variable $i$ is assigned to a value $j$ iff variable $j$ is assigned to value $i$. This constraint is useful in many real world problems in which a set of entities need to be paired up; particularly, in sport scheduling problems teams need to form a set of pairs that define the games.

A `symmetric_alldifferent` achieving domain consistency provides more pruning power than the equivalent decomposition given by the `alldifferent` constraint and the set of $x_i = j \iff x_j = i$ constraints (Régin, 1999). Its filtering algorithm is inspired from the one for `alldifferent` with the difference being that the matching is computed in a graph (not necessarily bipartite) called *contracted value graph* where vertices and values representing the same entity are collapsed into a single vertex (i.e. the vertex $x_i$ and the vertex $i$ are merged into a single vertex $i$ representing both the variable and the value). Régin proved that there is a bijection between a matching in the contracted value graph and a solution of the `symmetric_alldifferent` constraint. Therefore, counting the number of matchings on the contracted value graph corresponds to counting the number of solutions to the constraint.

Friedland (2008) and Alon and Friedland (2008) extended the Brégman-Minc upper bound to consider the number of matchings in general undirected graphs. Therefore, we can exploit the bound as in the previous section in order to provide an upper bound of the solution count and the solution densities for the `symmetric_alldifferent` constraint. The upper bound for the number of matchings of a graph $G = (V, E)$ representing the contracted value graph is the following:

$$\#matchings(G) \leq \prod_{v \in V} (deg(v))!^{\frac{1}{2deg(v)}} \tag{4}$$

where $deg(v)$ is the degree of the vertex $v$ and $\#matchings(G)$ denotes the number of matchings on the graph $G$. Note that in case of a bipartite graph, this bound is equivalent to the Brégman-Minc upper bound.

The algorithm for counting the number of solutions and computing the solution densities can be easily derived from what we proposed for the `alldifferent`.

**Example 1.** *Consider a* `symmetric_alldifferent` *defined on six variables* $x_1, \ldots, x_6$ *each one having a domain equal to* $\{1, \ldots, 6\}$*. In Figure 1 the associated contracted value graph is depicted (together with a possible solution to the constraint). In this case, the number of solutions of the* `symmetric_alldifferent` *can be computed as* $5 * 3 = 15$*. In the contracted value graph each vertex is connected to each other vertex, forming a clique of size 6, therefore all the vertices have a degree equal to 5. The upper bound proposed by Friedland is equal to:*

$$\#matchings(G) \leq \prod_{v \in V} (deg(v))!^{\frac{1}{2deg(v)}} = (5!^{1/10})^6 \approx 17.68$$

*In the* `alldifferent` *formulation, the related value graph has variable vertices connected to each of the values (from 1 to 6) thus the* $r_i$*'s are equal to 6. If we consider to rule out all the edges causing degenerated assignments (*$x_i = i$*) then we end up with a value graph in which all the* $r_i$*'s are equal to 5. The Brégman-Minc upper bound would give:*

$$perm(A) \leq \prod_{i=1}^{n} (r_i!)^{1/r_i} = (5!^{(1/5)})^6 \approx 312.62.$$

*The result is obviously very far from the upper bound given by Formula 4 as well as from the exact value.*

## 4. Counting for Global Cardinality Constraints

We present in this section how to extend the results obtained in Section 3 to the Global Cardinality Constraint (`gcc`), which is a generalization of the `alldifferent` constraint.

**Definition 5** (Global Cardinality Constraint)**.** *The set of solutions of constraint* $\texttt{gcc}(X, l, u)$ *where* $X$ *is a set of* $k$ *variables,* $l$ *and* $u$ *respectively the lower and upper bounds for each value, is defined as:*

$$T(gcc(X, l, u)) = \{(d_1, \ldots, d_k) \mid d_i \in D_i, l_d \leq |\{d_i | d_i = d\}| \leq u_d \, \forall d \in D_X = \bigcup_{x_j \in X} D_j\}$$

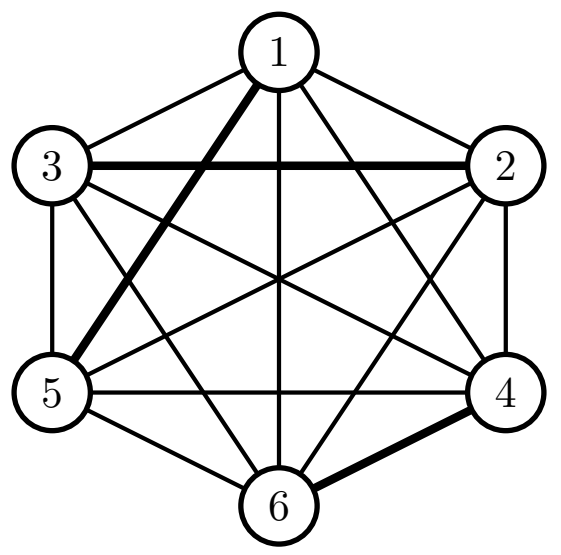

Figure 1: Contracted Value Graph for the constraint symmetric_alldifferent of Example 1. Edges in bold represent a possible solution.

We will consider a gcc in which all the fixed variables are removed and the lower and upper bounds are adjusted accordingly (the semantics of the constraint is unchanged). We refer to the new set of variables as $X' = \{x \in X \mid x \text{ is not bound}\}$; lower bounds are $l'$ where $l'_d = l_d - |\{x \in X \mid x = d\}|$ and upper bounds $u'$ are defined similarly; we assume the constraint maintains $\theta$-consistency so $l'_d \geq 0$ and $u'_d \geq 0$ for each $d \in D_X$.

Inspired by Quimper, Lopez-Ortiz, van Beek, and Golynski (2004) and Zanarini, Milano, and Pesant (2006), we define $G_l$ the lower bound graph.

**Definition 6.** *Let $G_l(X' \cup D_l, E_l)$ be an undirected bipartite graph where $X'$ is the set of unbounded variables and $D_l$ the extended value set, that is for each $d \in D_X$ the graph has $l'_d$ vertices $d^1, d^2, \ldots$ representing $d$ ($l'_d$ possibly equal to zero). There is an edge $(x_i, d^j) \in E_l$ if and only if $d \in D_i$.*

Note that a maximum matching on $G_l$ corresponds to a partial assignment of the variables in $X$ that satisfies the gcc lower bound restriction on the number of occurrences of each value. This partial assignment may or may not be completed to a full assignment that satisfies both upper bound and lower bound restrictions (here we do not take into consideration augmenting paths as Zanarini et al., 2006 but instead we fix the variables to the values represented by the matching in $G_l$).

**Example 2.** *Suppose we have a* gcc *defined on $X = \{x_1, \ldots, x_6\}$ with domains $D_1 = D_4 = \{1, 2, 3\}$, $D_2 = \{2\}$, $D_3 = D_5 = \{1, 2\}$ and $D_6 = \{1, 3\}$; lower and upper bounds for the values are respectively $l_1 = 1$, $l_2 = 3$, $l_3 = 0$ and $u_1 = 2$, $u_2 = 3$, $u_3 = 2$. Considering that $x_2 = 2$, the lower and upper bounds for the value 2 are respectively $l'_2 = 2$ and $u'_2 = 2$. The lower bound graph is shown in Figure 2a: variable $x_2$ is bounded and thus does not appear in the graph, value vertex 2 is represented by two vertices because it has $l'_2 = 2$ (although $l_2 = 3$); finally value vertex 3 does not appear because it has a lower bound equal to zero. The matching shown in the figure (bold edges) is maximum. However if we fix the assignments represented by it ($x_1 = 2$, $x_4 = 2$, $x_6 = 1$) it is not possible to have a consistent solution since both $x_3$ and $x_5$ have to be assigned either to 1 or 2 hence exceeding the upper bound restriction. To compute the permanent two additional fake value vertices would be added to the graph and connected to all the variable vertices (not shown in the figure).*

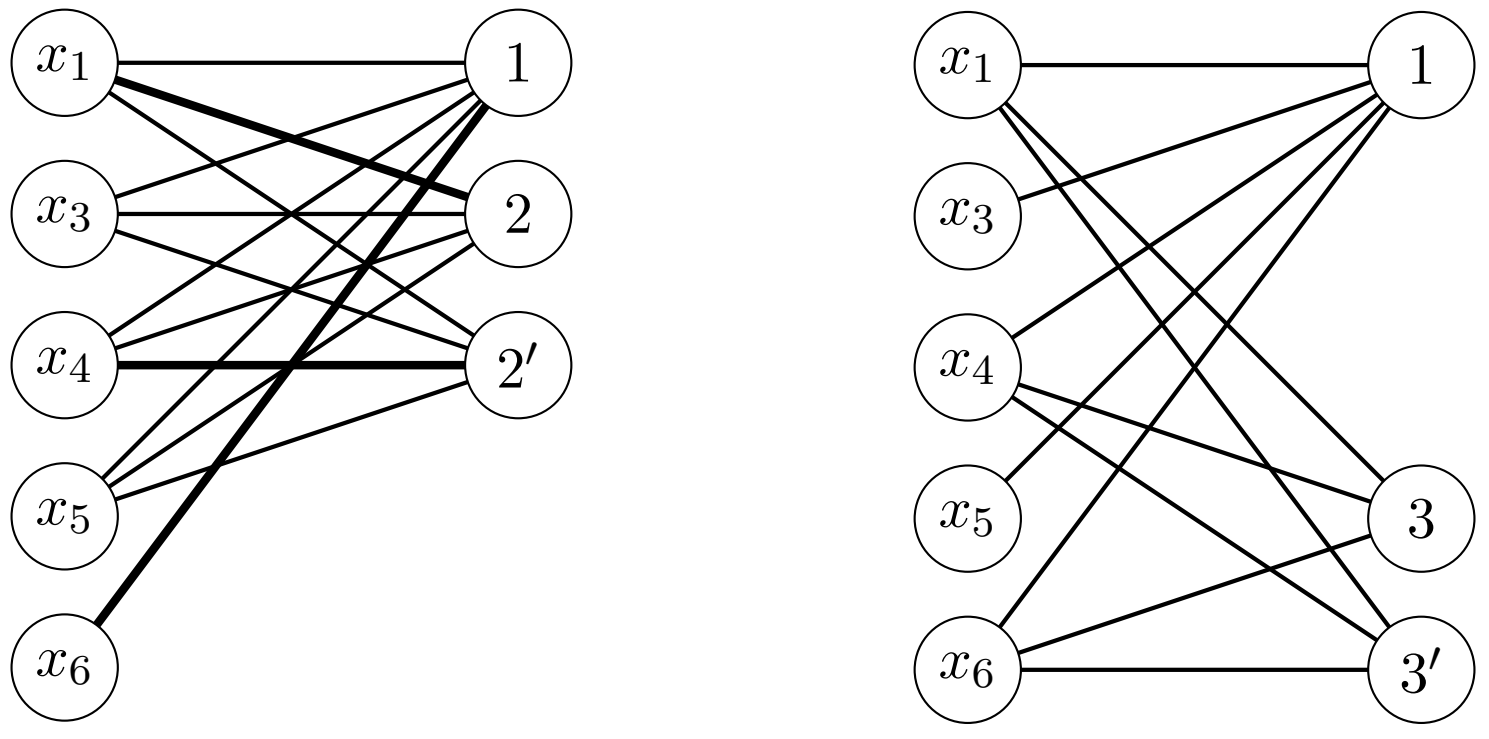

(a) (b)

Figure 2: Lower Bound Graph (a) and Residual Upper Bound Graph (b) for Example 2

Every partial assignment that satisfies just the lower bound restriction might correspond to several maximum matchings in $G_l$ due to the duplicated vertices. For each partial assignment satisfying the lower bound restriction there are exactly $\prod_{d \in D_X} l'_d!$ maximum matchings corresponding to that particular partial assignment. If we take into consideration Example 2 shown in Figure 2a, variables $x_1$ and $x_4$ may be matched respectively to any permutation of the vertices 2 and $2'$, however no matter which is the permutation, this set of matchings represents always the assignment of both $x_2$ and $x_4$ to the value 2.

Let $M_l$[5] be the set of maximum matchings in $G_l$. We define $f : M_l \rightarrow \mathbb{N}$, a function that counts the number of possible ways a maximum matching can be extended to a full `gcc` solution. As shown in Example 2, $f$ can be possibly equal to zero. Note that the number of the remaining variables that need to be assigned starting from a matching $m \in M_l$ is equal to $K = |X'| - \sum_{d \in D_X} l'_d$.

The total number of solutions satisfying the `gcc` is:

$$\#gcc(X, l, u) = \frac{\sum_{m \in M_l} f(m)}{\prod_{d \in D_X} l'_d!} \leq \frac{|M_l| \max_{m \in M_l}(f(m))}{\prod_{d \in D_X} l'_d!} \leq \frac{UB(G_l) \max_{m \in M_l}(f(m))}{\prod_{d \in D_X} l'_d!} \tag{5}$$

where $UB(G_l)$ represents an upper bound on the permanent of the $0-1$ matrix corresponding to graph $G_l$.

Note that computing $f(m)$ is as hard as computing the permanent. In fact if $l$ and $u$ are respectively equal to 0 and 1 for each value, the result is an `alldifferent` constraint and equation 5 simplifies to $\#gcc(X, l, u) = f(m)$ where $m = \{\emptyset\}$ and $f(m)$ corresponds to the permanent.

As computing $f(m)$ is a #P-complete problem on its own, we focus here on upper bounding $f(m)$. In order to do that, we introduce the upper bound residual graph. Intuitively, it is similar to the lower bound graph but it considers the upper bound restriction.

**Definition 7.** *Let $G_u(X' \cup D_u, E_u)$ be an undirected bipartite graph where $X'$ is the set of unbounded variables and $D_u$ the extended value set, that is for each $d \in D_X$ the graph has*

5. if $\forall d \in D_X, l'_d = 0$ then $M_l = \{\emptyset\}$ and $|M_l| = 1$

*$u'_d - l'_d$ vertices $d^1, d^2, \ldots$ representing $d$ (if $u'_d - l'_d$ is equal to zero then there is no vertex representing $d$). There is an edge $(x_i, d^j) \in E_u$ if and only if $d \in D_i$ and $u'_d - l'_d > 0$.*

Similarly to the lower bound matching, a matching on $G_u$ that covers $K$ variables may or may not be completed to a full assignment satisfying the complete gcc. Figure 2b shows the residual upper bound graph for Example 1: value 2 disappears from the graph since it has $u'_2 = l'_2$ i.e. starting from a matching in the lower bound graph, the constraints on value 2 are already satisfied.

In order to compute $\max_{m \in M_l}(f(m))$, we should build $\binom{|X|}{K}$ graphs each with a combination of $K$ variables, and then choose the one that maximizes the permanent. More practically, given the nature of the $UB^{MB}$ and $UB^{LB}$, it suffices to choose $K$ variables which contribute with the highest factor in the computation of the upper bounds; this can be easily done in $O(n \log K)$ by iterating over the $n$ variables and maintaining a heap with $K$ entries with the highest factor. We write $\hat{G}_u$ and $\tilde{G}_u$ for the graphs in which only the $K$ variables that maximize respectively $UB^{MB}$ and $UB^{LB}$ are present; note that $\hat{G}_u$ might be different from $\tilde{G}_u$.

We recall here that although only $K$ variables are chosen, the graphs $\hat{G}_u$ and $\tilde{G}_u$ are completed with fake vertices in such a way to have an equal number of vertices on the two vertex partitions. As in the lower bound graph, the given upper bound has to be scaled down by a factor of $\prod_{d \in D_X}(u'_d - l'_d)!$. From Equation 5, the number of gcc solutions is bounded from above by:

$$\#gcc(X, l, u) \leq \frac{UB(G_l) \min(UB^{MB}(\hat{G}_u), UB^{LB}(\tilde{G}_u))}{\prod_{d \in D_X}(l'_d!(u'_d - l'_d)!)} \tag{6}$$

Scaling and also fake vertices used with the permanent bounds are factors that degrade the quality of the upper bound. Nonetheless, solution densities are computed as a ratio between two upper bounds therefore these scaling factors are often attenuated.

**Example 3.** *We refer to the* gcc *described in Example 2. The exact number of solutions is 19. The $UB^{MB}$ and $UB^{LB}$ for the lower bound graph in Figure 2a are both* 35 *(the scaling for the two fake value vertices is already considered). In the upper bound only 2 variables need to be assigned and the one maximizing the bounds are $x_1$ and $x_4$ (or possibly $x_6$): the resulting permanent upper bound is* 6*. An upper bound on the total number of* gcc *solutions is then $\lfloor \frac{35*6}{4} \rfloor = 52$ where the division by 4 is due to $l'_2! = 2!$ and $u'_3! = 2!$.*

*Figure 3 shows the lower bound and residual upper bound graph for the same constraint where $x_1 = 1$ and domain consistency is achieved. Vertex $x_1$ has been removed and $l'_1 = 0$ and $u'_1 = 1$. The graph $G_l$ has a permanent upper bound of 6. The number of unassigned variables in $G_u$ is 2 and the ones maximizing the upper bounds are $x_4$ and $x_6$, giving an upper bound of 6. The total number of* gcc *solutions with $x_1 = 1$ is then bounded above by $\lfloor \frac{6*6}{4} \rfloor = 9$; the approximate solution density before normalizing it is thus $9/52$. Note that after normalization, it turns out to be about $0.18$ whereas the exact computation of it is $5/19 \sim 0.26$.*

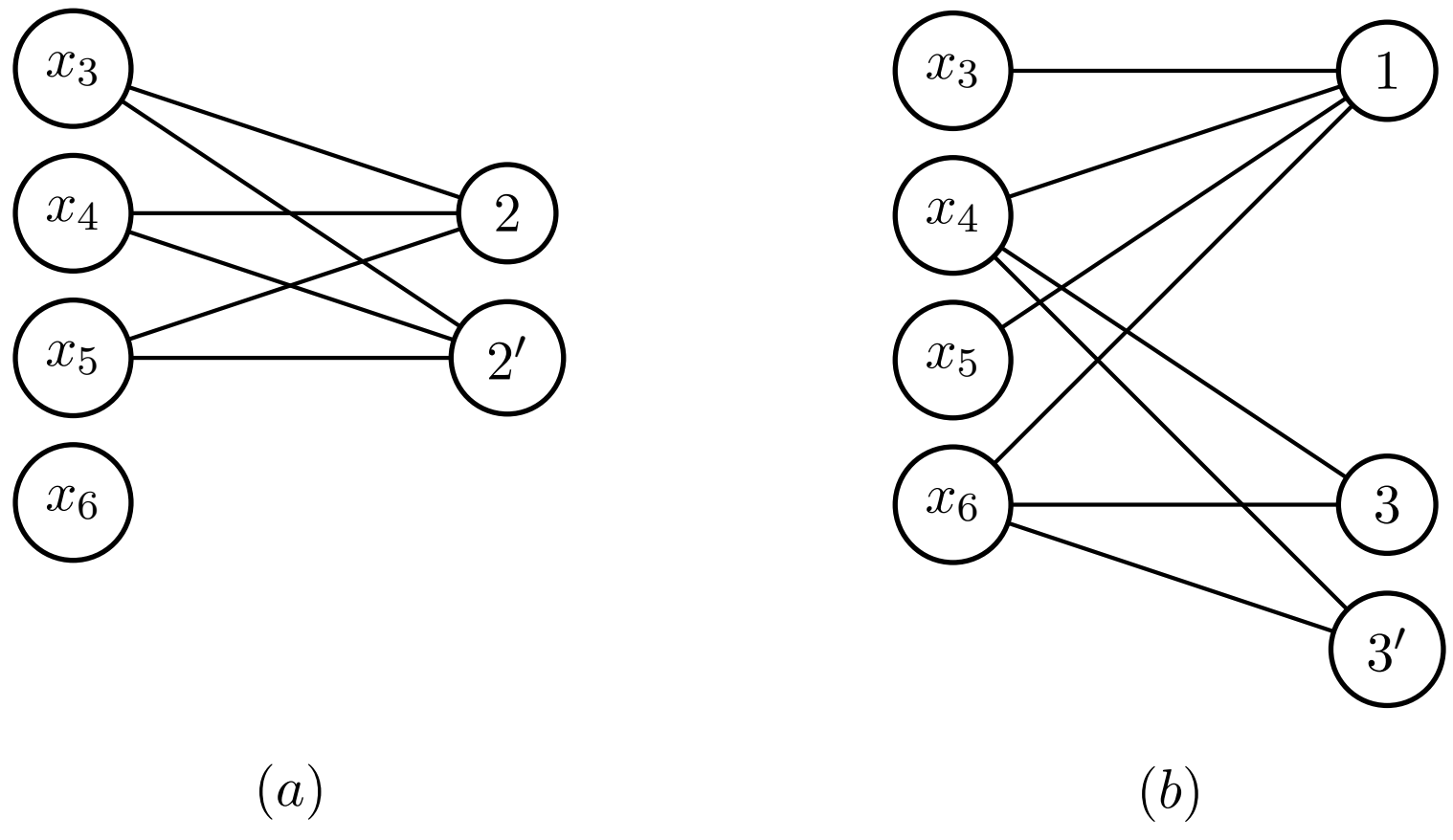

Figure 3: Lower Bound Graph (a) and Residual Upper Bound Graph (b) assuming $x_1 = 1$

## 5. Counting for Regular and Knapsack Constraints

The `regular` constraint is useful to express patterns that must be exhibited by sequences of variables.

**Definition 8** (Regular Language Membership Constraint). *The* `regular`$(X, \Pi)$ *constraint holds if the values taken by the sequence of finite domain variables* $X = \langle x_1, x_2, \ldots, x_k \rangle$ *spell out a word belonging to the regular language defined by the deterministic finite automaton* $\Pi = (Q, \Sigma, \delta, q_0, F)$ *where* $Q$ *is a finite set of states,* $\Sigma$ *is an alphabet,* $\delta : Q \times \Sigma \to Q$ *is a partial transition function,* $q_0 \in Q$ *is the initial state, and* $F \subseteq Q$ *is the set of final (or accepting) states.*

Linear equalities and inequalities are expressed as `knapsack` constraints.

**Definition 9** (Knapsack Constraint). *The* `knapsack`$(\mathbf{x}, \mathbf{c}, \ell, u)$ *constraint holds if*

$$\ell \le \mathbf{cx} \le u$$

*where* $\mathbf{c} = (c_1, c_2, \ldots, c_k)$ *is an integer row vector,* $\mathbf{x}$ *is a column vector of finite domain variables* $(x_1, x_2, \ldots, x_k)^T$ *with* $x_i \in D_i$*, and* $\ell$ *and* $u$ *are integers.*

We assume that $l$ and $u$ are finite as they can always be set to the smallest and largest value that $\mathbf{cx}$ can take. Strictly speaking to be interpreted as a knapsack, the integer values involved (including those in the finite domains) should be nonnegative but the algorithms proposed in this section can be easily adapted to lift the restriction of nonnegative coefficients and domain values, at the expense of a larger graph in the case of the algorithm of Section 5.1. So we are dealing here with general linear constraints.

The filtering algorithms for the `regular` constraint and the `knapsack` constraint (when domain consistency is enforced) are both based on the computation of paths in a layered acyclic directed graph (Pesant, 2004; Trick, 2003). This graph has the property that paths from the first layer to the last are in one-to-one correspondence with solutions of the constraint. An exact counting algorithm for the former constraint is derived by Zanarini and Pesant (2009) — in the next section we describe an exact counting algorithm for `knapsack` constraints which is similar in spirit, while in Section 5.2 we present an approximate counting algorithm attuned to bounds consistency. [6]

### 5.1 Domain Consistent Knapsacks

We start from the reduced graph described by Trick (2003), which is a layered directed graph $G(V,A)$ with special vertex $v_{0,0}$ and a vertex $v_{i,b} \in V$ for $1 \le i \le k$ and $0 \le b \le u$ whenever

$$\forall\, j \in [1,i],\ \exists\, d_j \in D_j \text{ such that } \sum_{j=1}^{i} c_j d_j = b$$

and

$$\forall\, j \in (i,n],\ \exists\, d_j \in D_j \text{ such that } \ell - b \le \sum_{j=i+1}^{k} c_j d_j \le u - b,$$

and an arc $(v_{i,b}, v_{i+1,b'}) \in A$ whenever

$$\exists\, d \in D_{i+1} \text{ such that } c_{i+1} d = b' - b.$$

We define the following two recursions to represent the number of incoming and outgoing paths at each node.

For every vertex $v_{i,b} \in V$, let $\#ip(i,b)$ denote the number of paths from vertex $v_{0,0}$ to $v_{i,b}$:

$$\begin{aligned} \#ip(0,0) &= 1 \\ \#ip(i+1,b') &= \sum_{(v_{i,b},v_{i+1,b'})\in A} \#ip(i,b), \quad 0 \le i < n \end{aligned}$$

Let $\#op(i,b)$ denote the number of paths from vertex $v_{i,b}$ to a vertex $v_{k,b'}$ with $\ell \le b' \le u$.

$$\begin{aligned} \#op(n,b) &= 1 \\ \#op(i,b) &= \sum_{(v_{i,b},v_{i+1,b'})\in A} \#op(i+1,b'), \quad 0 \le i < k \end{aligned}$$

The total number of paths (i.e. the *solution count*) is given by

$$\#\texttt{knapsack}(\mathbf{x},\mathbf{c},\ell,u) = \#op(0,0)$$

in time linear in the size of the graph even though there may be exponentially many of them. The *solution density* of variable-value pair $(x_i, d)$ is given by

$$\sigma(x_i, d, \texttt{knapsack}) = \frac{\sum_{(v_{i-1,b},v_{i,b+c_i d})\in A} \#ip(i-1,b) \cdot \#op(i, b+c_i d)}{\#op(0,0)}.$$

6. This was originally introduced by Pesant and Quimper (2008).

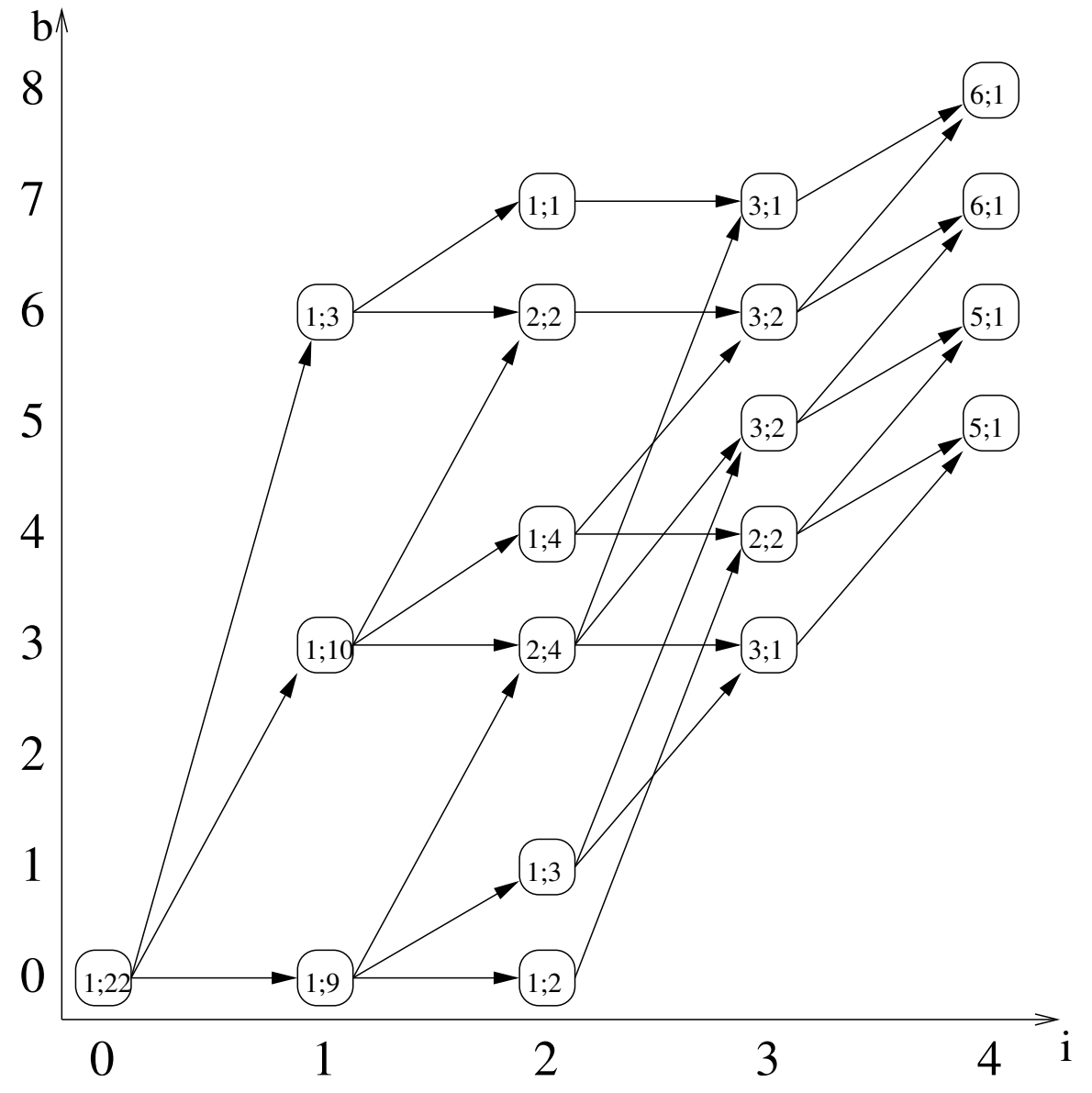

Figure 4: Reduced graph for knapsack constraint $5 \leq 3x_1 + x_2 + 2x_3 + x_4 \leq 8$ with $D_1 = \{0,1,2\}, D_2 = \{0,1,3\}, D_3 = \{0,1,2\}, D_4 = \{1,2\}$. Vertex labels represent the number of incoming and outgoing paths.

| | variable | | | |
|---|---|---|---|---|
| value | $x_1$ | $x_2$ | $x_3$ | $x_4$ |
| 0 | 9/22 | 8/22 | 9/22 | – |
| 1 | 10/22 | 8/22 | 7/22 | 11/22 |
| 2 | 3/22 | – | 6/22 | 11/22 |
| 3 | – | 6/22 | – | – |

Table 1: Solution densities for the example of Fig. 4.

In Figure 4, the left and right labels inside each vertex give the number of incoming and outgoing paths for that vertex, respectively. Table 1 reports the solution densities for every variable-value pair.

The time required to compute recursions $\#ip()$ and $\#op()$ is related to the number of arcs, which is in $\mathcal{O}(ku \max_{1 \leq i \leq k}\{|D_i|\})$. Then each solution density computes a summation over a subset of the arcs but each arc of the graph is involved in at most one such summation, so the overall time complexity of computing every solution density is $\mathcal{O}(ku \max_{1 \leq i \leq k}\{|D_i|\})$ as well.

### 5.2 Bounds Consistent Knapsacks

Knapsack constraints, indeed most arithmetic constraints, have traditionally been handled by enforcing bounds consistency, a much cheaper form of inference. In some situations, we may not afford to enforce domain consistency in order to get the solution counting information we need to guide our search heuristic. Can we still retrieve such information, perhaps not as accurately, from the weaker bounds consistency?

Consider the variable $x$ with domain $D = [a, b]$. Each value in $D$ is equiprobable. We associate to $x$ the discrete random variable $X$ which follows a discrete uniform distribution with probability mass function $f(v)$, mean $\mu = E[X]$, and variance $\sigma^2 = Var[X]$.

$$f(v) = \begin{cases} \frac{1}{b-a+1} & \text{if } a \leq v \leq b \\ 0 & \text{otherwise} \end{cases} \tag{7}$$

$$\mu = \frac{a+b}{2} \tag{8}$$

$$\sigma^2 = \frac{(b-a+1)^2 - 1}{12} \tag{9}$$

To find the distribution of a variable subject to a knapsack constraint, one needs to find the distribution of a linear combination of uniformly distributed random variables. Lyapunov's central limit theorem allows us to approximate the distribution of such a linear combination.

**Theorem 1** (Lyapunov's central limit theorem)**.** *Consider the independent random variables $X_1, \ldots, X_n$. Let $\mu_i$ be the mean of $X_i$, $\sigma_i^2$ be its variance, and $r_i^3 = E[|X_i - \mu_i|^3]$ be its third central moment. If*

$$\lim_{n \to \infty} \frac{(\sum_{i=1}^n r_i^3)^{\frac{1}{3}}}{(\sum_{i=1}^n \sigma_i^2)^{\frac{1}{2}}} = 0,$$

*then the random variable $S = \sum_{i=1}^n X_i$ follows a normal distribution with mean $\mu_S = \sum_{i=1}^n \mu_i$ and variance $\sigma_S^2 = \sum_{i=1}^n \sigma_i^2$.*

The probability mass function of the normal distribution with mean $\mu$ and variance $\sigma^2$ is the Gaussian function:

$$\varphi(x) = \frac{e^{-\frac{(x-\mu)^2}{2\sigma^2}}}{\sigma\sqrt{2\pi}} \tag{10}$$

Note that Lyapunov's central limit theorem does not assume that the variables are taken from identical distributions. This is necessary since variables with different domains have different distributions.

Lemma 1 defines an upper bound on the third central moment of the expression $kX$ where $k$ is a positive coefficient and $X$ is a uniformly distributed random variable.

**Lemma 1.** *Let $Y$ be a discrete random variable equal to $kX$ such that $k$ is a positive coefficient and $X$ is a discrete random variable uniformly distributed over the interval $[a, b]$. The third central moment $r^3 = E[|Y - E[Y]|^3]$ is no greater than $k^3(b-a)^3$.*

*Proof.* The case where $a = b$ is trivial. We prove for $b - a > 0$. The proof involves simple algebraic manipulations from the definition of the expectation.

$$
\begin{aligned}
r^3 &= \sum_{i=ka}^{kb} |i - E[Y]|^3 f(i) && (11)\\
&= \sum_{j=a}^{b} |kj - kE[X]|^3 f(j) && (12)\\
&= k^3 \sum_{j=a}^{b} \left| j - \frac{a+b}{2} \right|^3 \frac{1}{b-a+1} \text{ since } k > 0 && (13)\\
&= \frac{k^3}{b-a+1} \left( \sum_{j=a}^{\frac{a+b}{2}} \left( \frac{a+b}{2} - j \right)^3 + \sum_{j=\frac{a+b}{2}}^{b} \left( j - \frac{a+b}{2} \right)^3 \right) && (14)\\
&= \frac{k^3}{b-a+1} \left( \sum_{j=0}^{\frac{b-a}{2}} j^3 + \sum_{j=0}^{\frac{b-a}{2}} j^3 \right) && (15)\\
&\leq \frac{2k^3}{b-a} \sum_{j=0}^{\frac{b-a}{2}} j^3 \text{ since } b - a > 0 && (16)
\end{aligned}
$$

Let $m = \frac{b-a}{2}$.

$$
\begin{aligned}
r^3 &\leq \frac{k^3}{m}\sum_{j=0}^{m} j^3 &(17)\\
&\leq \frac{k^3}{m}\left(\frac{1}{4}(m+1)^4 - \frac{1}{2}(m+1)^3 + \frac{1}{4}(m+1)^2\right) &(18)\\
&\leq \frac{k^3}{m}\left(\frac{m^4}{4} + \frac{m^3}{2} + \frac{m^2}{4}\right) &(19)\\
&\leq \frac{k^3}{m}\left(\frac{m^4}{4} + m^4 + m^4\right) \text{ since } m \geq \tfrac{1}{2} &(20)\\
&\leq \frac{9}{4}k^3 m^3 &(21)
\end{aligned}
$$

Which confirms that $r^3 \leq \frac{9}{32}k^3(b-a)^3 \leq k^3(b-a)^3$. □

Lemma 2 defines the distribution of a linear combination of uniformly distributed random variables.

**Lemma 2.** *Let $Y = \sum_{i=1}^{n} c_i X_i$ be a random variable where $X_i$ is a discrete random variable uniformly chosen from the interval $[a_i, b_i]$ and $c_i$ is a non-negative coefficient. When $n$ tends to infinity, the distribution of $Y$ tends to a normal distribution with mean $\sum_{i=1}^{n} c_i \frac{a_i+b_i}{2}$ and variance $\sum_{i=1}^{n} c_i^2 \frac{(b_i-a_i+1)^2-1}{12}$.*

*Proof.* Let $Y_i = c_i X_i$ be a random variable. We want to characterize the distribution of $\sum_{i=1}^{n} Y_i$. Let $m_i = \frac{b_i-a_i}{2}$. The variance of the uniform distribution over the interval $[a_i, b_i]$ is $\sigma_i^2 = \frac{(b_i-a_i+1)^2-1}{12} = \frac{(m_i+\frac{1}{2})^2}{3} - \frac{1}{12}$. We have $Var[Y_i] = c_i^2 Var[X_i] = c_i^2\sigma_i^2$. Let $r_i^3$ be the third central moment of $Y_i$. By Lemma 1, we have $r_i^3 \leq c_i^3(b_i-a_i)^3$. Let $L$ be the term mentioned in the condition of Lyapunov's central limit theorem:

$$
L = \lim_{n\to\infty} \frac{\left(\sum_{i=1}^{n} r_i^3\right)^{\frac{1}{3}}}{\left(\sum_{i=1}^{n} c_i^2\sigma_i^2\right)^{\frac{1}{2}}} \qquad (22)
$$

Note that the numerator and the denominator of the fraction are non-negative. This implies that $L$ itself is non-negative. We prove that $L \leq 0$ as $n$ tends to infinity.

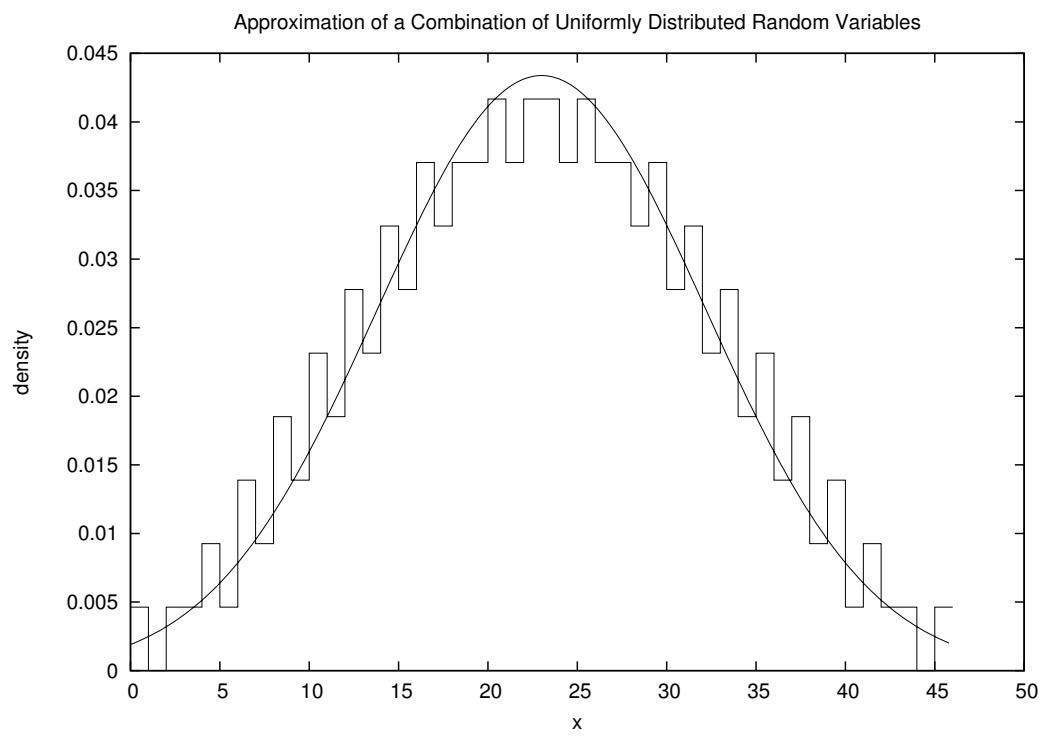

Figure 5: The histogram is the actual distribution of the expression $3x+4y+2z$ for $x, y, z \in [0, 5]$. The curve is the approximation given by the Gaussian curve with mean $\mu = 22.5$ and variance $\sigma^2 = 84.583$.

$$
\begin{aligned}
L &\leq \lim_{n\to\infty} \frac{\left(\sum_{i=1}^{n} 8c_i^3 m_i^3\right)^{\frac{1}{3}}}{\left(\sum_{i=1}^{n} c_i^2 \left(\frac{(m_i+\frac{1}{2})^2}{3} - \frac{1}{12}\right)\right)^{\frac{1}{2}}} && (23)\\
&\leq \lim_{n\to\infty} \frac{\left(8\sum_{i=1}^{n} c_i^3 m_i^3\right)^{\frac{1}{3}}}{\left(\frac{1}{3}\sum_{i=1}^{n} c_i^2 m_i^2\right)^{\frac{1}{2}}} && (24)\\
&\leq \lim_{n\to\infty} 2\sqrt{3}\sqrt[6]{\frac{\left(\sum_{i=1}^{n} c_i^3 m_i^3\right)^2}{\left(\sum_{i=1}^{n} c_i^2 m_i^2\right)^3}} && (25)\\
&\leq \lim_{n\to\infty} 2\sqrt{3}\sqrt[6]{\frac{\sum_{i=1}^{n}\sum_{j=1}^{n}(c_i c_j m_i m_j)^3}{\sum_{i=1}^{n}\sum_{j=1}^{n}\sum_{k=1}^{n}(c_i c_j c_k m_i m_j m_k)^2}} && (26)
\end{aligned}
$$

Note that in the last inequality, the terms $(c_i c_j m_i m_j)^3$ and $(c_i c_j c_k m_i m_j m_k)^2$ are of the same order. However, there are $n$ times more terms in the denominator than the numerator. Therefore, when $n$ tends to infinity, the fraction tends to zero which proves that $L = 0$ as $n$ tends to zero.

By Lyapunov's central limit theorem, as $n$ tends to infinity, the expression $Y = \sum_{i=1}^{n} Y_i$ tends to a normal distribution with mean $E[Y] = \sum_{i=1}^{n} c_i E[X_i] = \sum_{i=1}^{n} c_i \frac{a_i+b_i}{2}$ and variance $Var[Y] = \sum_{i=1}^{n} c_i^2 Var[X_i] = \sum_{i=1}^{n} c_i^2 \frac{(b_i-a_i+1)^2-1}{12}$. □

Consider the knapsack constraint $\ell \leq \sum_{i=1}^{n} c_i x_i \leq u$. Let $x_{n+1}$ be a variable with domain $D_{n+1} = [\ell, u]$. We obtain $x_j = \frac{1}{c_j}(x_{n+1} - \sum_{i=1}^{j-1} c_i x_i - \sum_{i=j+1}^{n} c_i x_i)$. Some coefficients in this expression might be negative. They can be made positive by setting $c_i' = -c_i$ and $D_i' = [-\max(D_i), -\min(D_i)]$. When $n$ grows to infinity, the distribution of $x_j$ tends to a normal distribution as stated in Lemma 2. In practice, the normal distribution is a good estimation even for small values of $n$. Figure 5.2 shows the actual distribution of the expression $3x + 4y + 2z$ for $x, y, z \in [0, 5]$ and its approximation by a normal distribution.

Given a variable $x_i$ subject to a knapsack constraint, Algorithm 2 returns the assignment $x_i = k_i$ with the highest solution density. The *for loop* computes the average mean $\mu_j$ and the variance $\sigma_j^2$ of the uniform distribution associated to each variable $x_j$. Lines 4 and 5 compute the mean and the variance of the distribution of $x_{n+1} - \sum_{j=1}^{n} c_j x_j$ while Lines 6 and 7 compute the mean and the variance of $x_i = \frac{1}{c_i}(x_{n+1} - \sum_{j=1}^{i-1} c_j x_j - \sum_{j=i+1}^{n} c_j x_j)$. Since this normal distribution is symmetric and unimodal, the most likely value $k_i$ in the domain $D_i$ is the one closest to the mean $\mu_i$. The algorithm finds and returns this value as well as its density $d_i$. The density $d_i$ is computed using the normal distribution. Since the variable $x_i$ must be assigned to a value in its domain, the algorithm normalizes on Line 9 the distribution over the values in the interval $[\min(D_i), \max(D_i)]$.

1 **for** $j \in [1, n]$ **do**<br>
2 $\quad \mu_j \leftarrow \frac{\min(D_j)+\max(D_j)}{2}$;<br>
3 $\quad \sigma_j^2 \leftarrow \frac{(\max(D_j)-\min(D_j)+1)^2-1}{12}$;<br>
4 $M \leftarrow \frac{l+u}{2} - \sum_{j=1}^{n} c_j \mu_j$;<br>
5 $V \leftarrow \frac{(u-l+1)^2-1}{12} + \sum_{j=1}^{n} c_j^2 \sigma_j^2$;<br>
6 $m \leftarrow \frac{M+c_i\mu_i}{c_i}$;<br>
7 $v \leftarrow \frac{V-c_i^2\sigma_i^2}{c_i^2}$;<br>
8 $k_i \leftarrow \arg\min_{k \in D_i} |k - m|$;<br>
9 $d_i \leftarrow e^{-\frac{(k_i-m)^2}{2v}} / \sum_{k=\min(D_i)}^{\max(D_i)} e^{-\frac{(k-m)^2}{2v}}$;<br>
10 **return** $\langle x_i = k_i, d_i \rangle$

**Algorithm 2**: $x_i = k_i$ with the highest density as well as its density $d_i$ for knapsack constraint $\ell \leq \sum_{i=1}^{n} c_i x_i \leq u$.

Lines 1 through 5 take $O(n)$ time to execute. Line 8 depends on the data structure used by the solver to encode a domain. We assume that the line takes $O(\log |D_i|)$ time to execute. The summation on Line 9 can be computed in constant time by approximating the summation with $\Phi_{m,v}(\max(D_i) + \frac{1}{2}) - \Phi_{m,v}(\min(D_i) + \frac{1}{2})$ where $\Phi_{m,v}$ is the normal cumulative distribution function with average $m$ and variance $v$. The constant $\frac{1}{2}$ is added for the continuity correction. Other lines have a constant running time. The total complexity of Algorithm 2 is therefore $O(n + \log |D_i|)$. Note that Line 1 to Line 5 do not depend on the value of $i$. Their computation can therefore be cached for subsequent calls to the function over the same knapsack constraint. Using this technique, finding the variable $x_i \in \{x_1, \ldots, x_n\}$ which has an assignment $x_i = k_i$ of maximum density takes $O(\sum_{i=1}^{n} \log |D_i|)$ time.

A source of alteration of the distribution are values in the interval which are absent from the actual domain. Bounds consistency approximates the domain of a variable with its smallest covering interval. In order to reduce the error introduced by this approximation, one can compute the actual mean and actual variance of a domain $D_i$ on Lines 2 and 3 instead of using the mean and the variance of the covering interval, at a revised overall cost of $O(\sum_{i=1}^{n} |D_i|)$.

## 6. Generic Constraint-Centered Counting-based Heuristics

The previous sections provided algorithms to retrieve solution counting information from many of the most frequently used constraints. That information must then be exploited to guide search. The solving process alternates between propagating constraints to filter domains and branching by fixing a variable to a value in its domain. The crucial choice of variable and value is made through a search heuristic. We considered many search heuristics based on counting information, which we describe briefly in the next paragraph. We will experiment extensively with one of the most successful ones in Section 7, so we present it in more detail. In the following, we denote by $C(x_i)$ the set of constraints whose scope contains the variable $x_i$. All the heuristics proposed assume a lexicographical ordering as tie breaking. Counting information is gathered at a search tree node once a propagation fixed point is reached: it is recomputed only on constraints for which a change occurred to the domain of a variable within its scope, and otherwise cached information is reused. That cached counting information is stored in trailing data structures (also known as reversible data structures) so that it can be retrieved upon backtracking. The heuristics considered fall into four broad categories:

**Combined choice of variable and value** Those that select directly a variable-value pair without an explicit differentiation of variable and value ordering, based on the aggregation, through simple functions, of the counting information coming from different constraints. Such heuristics iterate over each variable-value pair, aggregating solution densities from the relevant constraints and selecting the pair exhibiting the maximum aggregated score. The type of aggregation used is e.g. the maximum, minimum, sum, or average. For instance:

- maxSD: $\max_{c \in C(x_i)}(\sigma(x_i, d, c))$ – selects the maximum of the solution densities.
- maxRelSD: $\max_{c \in C(x_i)}(\sigma(x_i, d, c) - (1/|D_i|))$ – selects the maximum of the solution densities subtracting the average solution density for that given variable (i.e. $1/|D_i|$). It smoothes out the inherent solution densities differences due to domain cardinalities (as also the following aggregation function).
- maxRelRatio: $\max_{c \in C(x_i)}(\frac{\sigma(x_i,d,c)}{(1/|D_i|)})$ – selects the maximum of the ratio between the solution density and the average solution density for that given variable.
- aAvgSD: $\frac{\sum_{c \in C(x_i)} \sigma(x_i,d,c)}{|C(x_i)|}$ – computes the arithmetic average of the solution densities.
- wSCAvg: $\frac{\sum_{c \in C(x_i)} (\#c\sigma(x_i,d,c))}{\sum_{c \in C(x_i)} \#c}$ – computes the average of the solution densities weighted by the constraints' solution count. The weights tend to favor branchings on variable-value pairs that keep a high percentage of solutions on constraints with a high solution count.

**Choice of constraint first** Those that focus first on a specific constraint (e.g. based on its solution count) and then select a variable-value pair (as before) among the variables in the preselected constraint's scope. For instance, minSCMaxSD first selects the constraint with the lowest number of solutions and then restricts the choice of variable to those involved in this constraint, choosing the variable-value pair with the highest solution density. The rationale behind this heuristic is that the constraint with the fewest solutions is probably among the hardest to satisfy.

**Restriction of variables** Those that preselect a subset of variables with minimum domain size and then choose among them the one with the best variable-value pair according to counting information.

**Choice of value only** Those using some other generic heuristic for variable selection and solution densities for value selection.

**Heuristic maxSD** The heuristic maxSD (Algorithm 3) simply iterates over all the variable-value pairs and chooses the one that has the highest density; assuming that the $\sigma(x_i, d, c)$ are precomputed, the complexity of the algorithm is $O(qm)$ where $q$ is the number of constraints and $m$ is the sum of the cardinalities of the variables' domains. Interestingly, such a heuristic likely selects a variable with a small domain, in keeping with the *fail-first principle*, since its values have on average a higher density compared to a variable with many values (consider that the average density of a value is $\sigma(x_i, d, c) = \frac{1}{|D_i|}$). Note that each constraint is considered individually.

max $= 0$;
**for** *each constraint* $c(x_1, \ldots, x_k)$ **do**
  **for** *each unbound variable* $x_i \in \{x_1, \ldots, x_k\}$ **do**
    **for** *each value* $d \in D_i$ **do**
      **if** $\sigma(x_i, d, c) >$ max **then**
        $(x^\star, d^\star) = (x_i, d)$;
        max $= \sigma(x_i, d, c)$;
return branching decision "$x^\star = d^\star$";

**Algorithm 3**: The Maximum Solution Density search heuristic (maxSD)

## 7. Experimental Analysis

We performed a thorough experimental analysis in order to evaluate the performance of the proposed heuristics on eight different problems.[7] All the problems expose sub-structures that can be encapsulated in global constraints for which counting algorithms are known. Counting-based heuristics are of no use for random problems as this class of problems do not expose any structure; nonetheless real-life problems usually do present structure therefore the performance of the heuristics proposed may have a positive impact in the quest to provide generic and efficient heuristics for structured problems. The problems on which we experimented have different structures and different constraints with possibly different

7. The instances we used are available at www.crt.umontreal.ca/∼quosseca/fichiers/20-JAIRbenchs.tar.gz.

arities interconnected in different ways; thus, they can be considered as good representatives of the variety of problems that may arise in real life.

### 7.1 Quasigroup Completion Problem with Holes (QWH)

Also referred to as the Latin Square problem, the QWH is defined on a $n \times n$ grid whose squares each contain an integer from 1 to $n$ such that each integer appears exactly once per row and column (problem 3 of the CSPLib maintained in Gent, Walsh, Hnich, & Miguel, 2009). The most common model uses a matrix of integer variables and an `alldifferent` constraint for each row and each column. So each constraint is defined on $n$ variables and is of the same type; each variable is involved in two constraints and has the same domain (disregarding the clues). This is a very homogeneous problem. We tested on the 40 hard instances used by Zanarini and Pesant (2009) with $n = 30$ and 42% of holes (corresponding to the phase transition), generated following Gomes and Shmoys (2002).

### 7.2 Magic Square Completion Problem

The magic square completion problem (problem 19 of CSPLib) is defined on a $n \times n$ grid and asks to fill the square with numbers from 1 to $n^2$ such that each row, each column and each main diagonal sums up to the same value. In order to make them harder, the problem instances have been partially prefilled (half of the instances have 10% of the variables set and the other half, 50% of the variables set). The 40 instances ($9 \times 9$) are taken from the work of Pesant and Quimper (2008). This problem is modeled with a matrix of integer variables, a single `alldifferent` constraint spanning over all the variables and a `knapsack` constraint for each row, column and main diagonal. The problem involves different constraints although the majority are equality `knapsack` with the same arity.

### 7.3 Nonograms

A Nonogram (problem 12 of CSPLib) is built on a rectangular $n \times m$ grid and requires filling in some of the squares in the unique feasible way according to some clues given on each row and column. As a reward, one gets a pretty monochromatic picture. Each individual clue indicates how many sequences of consecutive filled-in squares there are in the row (column), with their respective size in order of appearance. For example, "2 1 5" indicates that there are two consecutive filled-in squares, then an isolated one, and finally five consecutive ones. Each sequence is separated from the others by at least one blank square but we know little about their actual position in the row (column). Such clues can be modeled with `regular` constraints. This is a very homogeneous problem, with constraints of identical type defined over $m$ or $n$ variables, and with each (binary) variable involved in two constraints. These puzzles typically require some amount of search, despite the fact that domain consistency is maintained on each clue. We experimented with 180 instances[8] of sizes ranging from $16 \times 16$ to $32 \times 32$.

8. Instances taken from http://www.blindchicken.com/~ali/games/puzzles.html

### 7.4 Multi Dimensional Knapsack Problem

The Multi dimensional knapsack problem was originally proposed as an optimization problem by the OR community. We followed the same approach as Refalo (2004) in transforming the optimization problem into a feasibility problem by fixing the objective function to its optimal value, thereby introducing a 0-1 equality `knapsack` constraint. The other constraints are upper bounded `knapsack` constraints on the same variables. We tested on three different set of instances for a total of 25 instances: the first set corresponds to the six instances used by Refalo, the second set and the third set come from the OR-Library (Weish[1-13] from Shi, 1979; PB[1,2,4] and HP[1,2] from Freville & Plateau, 1990). The first instance set have $n$, that is the number of variables, ranging from 6 to 50 and $m$, that is the number of constraints, from 5 to 10; in the second and third instance set $n$ varies from 27 to 60 and $m$ from 2 to 5. The problem involves only one kind of constraint and, differently from the previous problem classes, all the constraints are posted on the same set of variables.

### 7.5 Market Split Problem

The market split problem was originally introduced by Cornuéjols and Dawande (1999) as a challenge to LP-based branch-and-bound approaches. There exists both a feasibility and optimization version. The feasibility problem consists of $m$ 0-1 equality `knapsack` constraints defined on the same set of $10(m-1)$ variables. Even small instances ($4 \leq m \leq 6$) are surprisingly hard to solve by standard means. We used the 10 instances tested by Pesant and Quimper (2008) that were generated by Wassermann (2007). The Market Split Problem shares some characteristics with the Multi Dimensional Knapsack problem: the constraints are of the same type and they are posted on the same set of variables.

### 7.6 Rostering Problem

The rostering problem was inspired by a rostering context. The objective is to schedule $n$ employees over a span of $n$ time periods. In each time period, $n-1$ tasks need to be accomplished and one employee out of the $n$ has a break. The tasks are fully ordered 1 to $n-1$; for each employee the schedule has to respect the following rules: two consecutive time periods have to be assigned to either two consecutive tasks (in no matter which order i.e. $(t, t+1)$ or $(t+1, t)$) or to the same task (i.e. $(t, t)$); an employee can have a break after no matter which task; after a break an employee cannot perform the task that precedes the task prior to the break (i.e. $(t, \mathit{break}, t-1)$ is not allowed). The problem is modeled with one `regular` constraint per row and one `alldifferent` constraint per column. We generated 2 sets of 30 instances with $n = 10$ each with 5% preset assignments and respectively 0% and 2.5% of values removed.

### 7.7 Cost-Constrained Rostering Problem

The cost-constrained rostering problem was borrowed from Pesant and Quimper (2008) and the 10 instances as well. It is inspired by a rostering problem where $m$ employees ($m = 4$) have to accomplish a set of tasks in a $n$-day schedule ($n = 25$). No employee can perform the same task as another employee on the same day (`alldifferent` constraint on each day). Moreover, there is an hourly cost for making someone work, which varies both across employees and days. For each employee, the total cost must be equal to a randomly generated value (equality knapsack constraint for each employee). Finally, each instance has about 10 forbidden shifts i.e. there are some days in which an employee cannot perform a given task. In the following, we refer to this problem also as KPRostering. This problem presents constraints of different types that have largely different arities.

### 7.8 Traveling Tournament Problem with Predefined Venues (TTPPV)

The TTPPV was introduced by Melo, Urrutia, and Ribeiro (2009) and consists of finding an optimal single round robin schedule for a sport event. Given a set of $n$ teams, each team has to play against every other team. In each game, a team is supposed to play either at home or away, however no team can play more than three consecutive times at home or away. The particularity of this problem resides on the venue of each game that is predefined, i.e. if team $a$ plays against $b$ it is already known whether the game is going to be held at $a$'s home or at $b$'s home. A TTPPV instance is said to be balanced if the number of home games and the number of away games differ by at most one for each team; otherwise it is referred to as non-balanced or random. The problem is modeled with one `alldifferent` and one `regular` constraint per row and one `alldifferent` constraint per column. The TTPPV was originally introduced as an optimization problem where the sum of the traveling distance of each team has to be minimized, however Melo et al. (2009) show that it is particularly difficult to find a single feasible solution employing traditional integer linear programming methods. Balanced instances of size 18 and 20 (the number of teams denotes the instance size) were taking from roughly 20 to 60 seconds to find a first feasible solution with Integer Linear Programming; non-balanced instances could take up to 5 minutes (or even time out after 2 hours of computation). Furthermore six non-balanced instances are infeasible but the ILP approach proposed by Melo et al. were unable to prove it. Hence, the feasibility version of this problem already represents a challenge.

For every problem (unless specified otherwise): domain consistency is maintained during search[9], the counting algorithm for the `alldifferent` constraint is UB-FC (upper bounds with forward checking as the consistency level enforced), the search tree is binary (i.e. $x_i = j \lor x_i \neq j$), and traversed depth-first. All tests were performed on a AMD Opteron 2.2GHz with 1GB and Ilog Solver 6.6; the heuristics that involve some sort of randomization (either in the heuristic itself or in the counting algorithms employed) have been run 10 times and the average of the results has been taken into account. We set a timeout of 20 minutes for all problems and heuristics. We present the results by plotting the percentage of solved instances against time or backtracks.

### 7.9 Comparing Counting-Based Search Heuristics

We first compare several of the proposed search heuristics based on counting with respect to how well they guide search, measured as the number of backtracks required to find a solution. The important issue of overall runtime will be addressed in the following sections.

9. Even for `knapsack` constraints, comparative experimental results on the same benchmark instances, originally reported by Pesant and Quimper (2008), indicated that maxSD performed better with domain consistency and the associated counting algorithm.

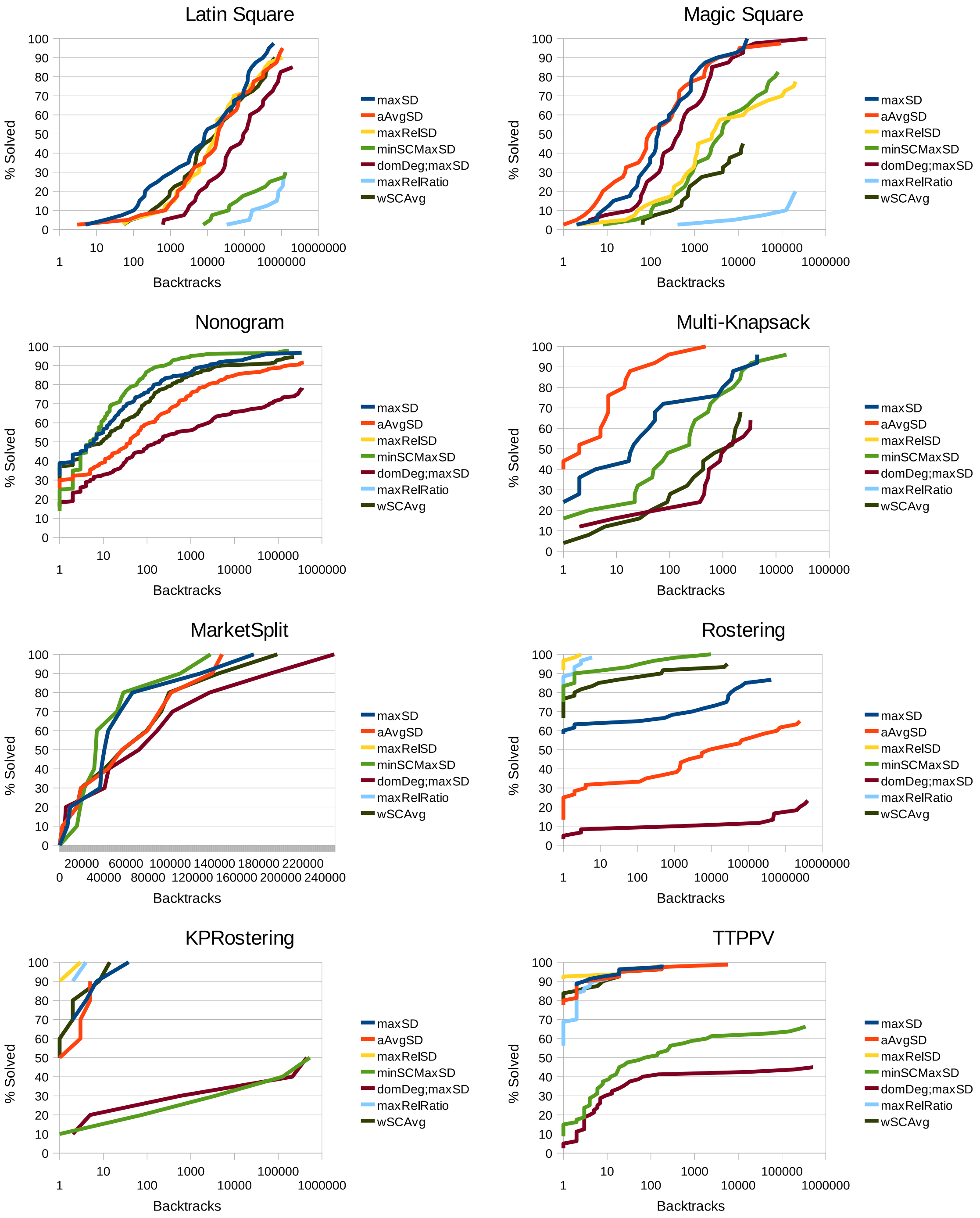

Figure 6: Percentage of solved instances with respect to the number of backtracks for the eight benchmark problems. The search heuristics compared are all based on solution counting.

Figure 6 plots the number of solved instances against backtracks for our eight benchmark problems. On the Nonogram, Multi-Knapsack, and Market Split problems, maxSD, maxRelSD, and maxRelRatio correspond to the same heuristics because domains are binary. Restricting the use of solution densities to the choice of a value once the variable has been selected by the popular domain size over dynamic degree heuristic (domDeg;maxSD) generally achieves very poor performance compared to the others. One disappointment which came as a surprise is that selecting first the constraint with the fewest solutions left (minSCMaxSD) often behaves poorly as well. For the Multi-Knapsack Problem aAvgSD, which takes the arithmetic average of the solution densities, performs about one order of magnitude better than the others. We believe that this might be explained by the fact that all the constraints share the same variables (in the Latin Square and Nonogram problems constraints overlap on only one variable): therefore branching while considering all the constraint information pays off. The maxSD and maxRelSD search heuristics stand out as being more robust on these benchmarks. They are quite similar but each performs significantly better than the other on one problem domain. Because it is slightly simpler, we will restrict ourselves to the former in the remaining experiments.

### 7.10 Comparing with Other Generic Search Heuristics

The experimental results of the previous section suggest that the relatively simple maxSD heuristic guides search at least as well as any of the others. We now compare it to the following ones (see Section 2 as a reference) which are good representatives of the state of the art for generic search heuristics:

- dom - it selects among the variables with smallest remaining domain uniformly at random and then chooses a value uniformly at random;
- domWDeg - it selects the variable according to the dom/wdeg heuristic and then the first value in lexicographic order;
- IBS - Impact-based Search with full initialization of the impacts; it chooses a subset of 5 variables with the best approximated impact and then it breaks ties based on the *node impacts* while further ties are broken randomly; (ILOG, 2005)

Figure 7 and 8 plot the number of solved instances against backtracks and time for our eight benchmark problems. For the moment we ignore the curves for the heuristics with restarts.

The maxSD heuristic significantly outperforms the other heuristics on the Latin Square, Magic Square, Multi Dimensional Knapsack, Cost-Constrained Rostering (KPRostering in the figure), and TTPPV problems (5 out of 8 problems), both in terms of number of backtracks and computation time. For the Nonogram Problem it is doing slightly worse than domWDeg and is eventually outperformed by IBS. The sharp improvement of the latter around 1000 backtracks suggests that singleton consistency is very powerful for this problem and not too time consuming since domains are binary. Indeed IBS's full initialization of the impacts at the root node achieves singleton consistency as a preprocessing step. This behavior is even more pronounced for the Rostering Problem (see the IBS curves). On that problem maxSD's performance is more easily compared to domWDeg, which dominates it.

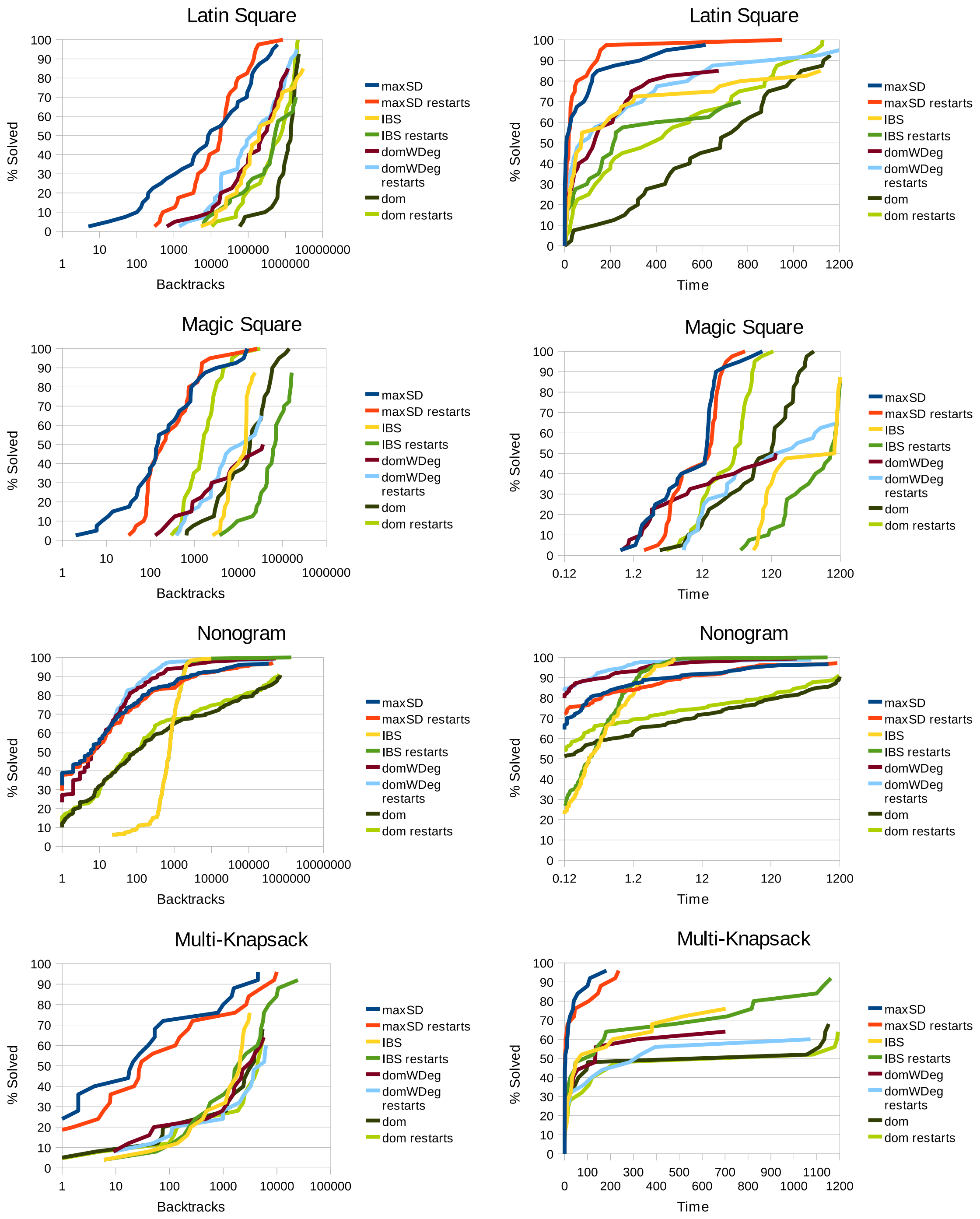

Figure 7: Percentage of solved instances with respect to the number of backtracks and to time (in seconds) for the first four benchmark problems. The search heuristics compared are maxSD, dom, IBS, and domWDeg, both with and without restarts.

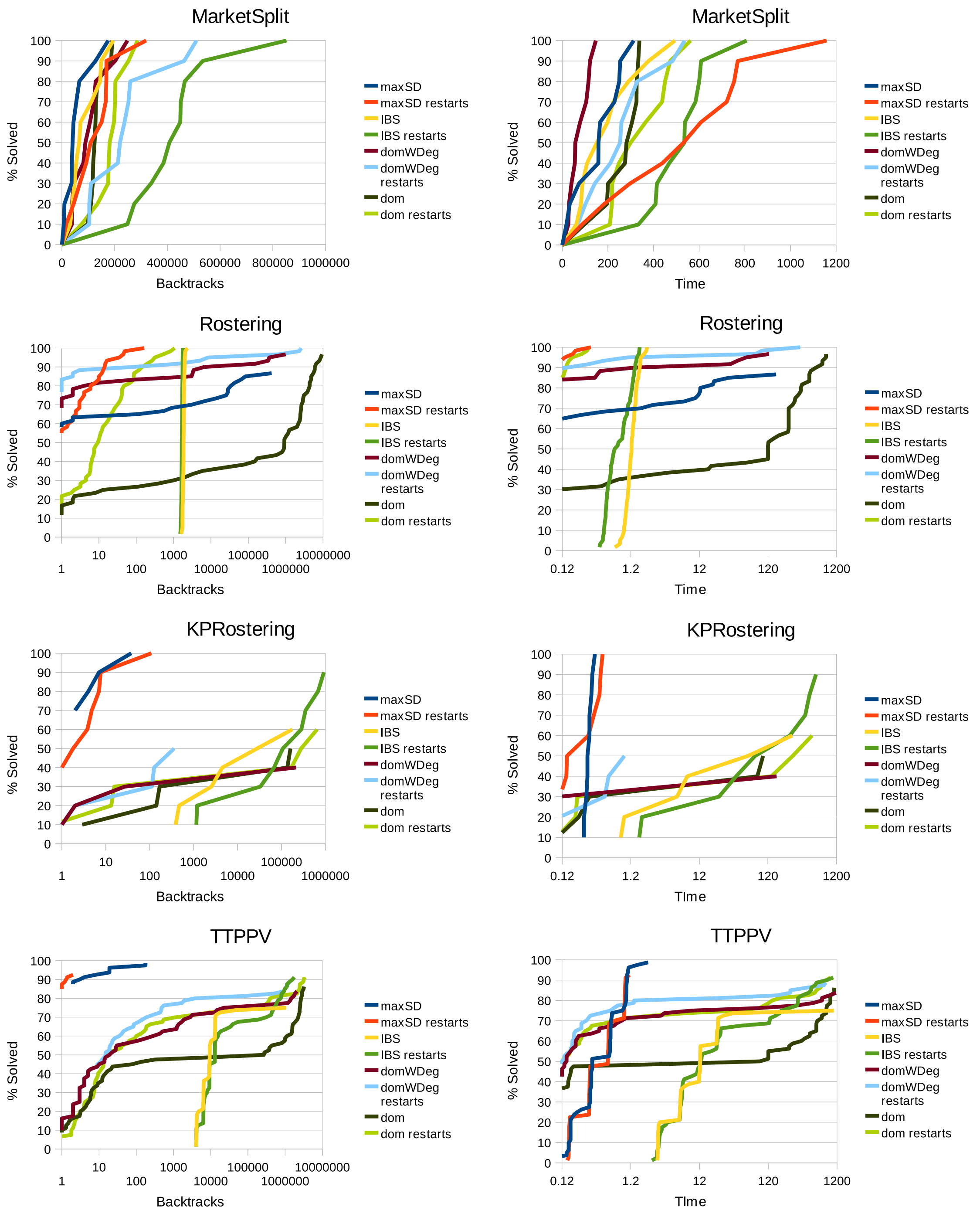

Figure 8: Percentage of solved instances with respect to the number of backtracks and to time (in seconds) for the last four benchmark problems. The search heuristics compared are maxSD, dom, IBS, and domWDeg, both with and without restarts.

For the Market Split Problem the differences in performance are not as striking: maxSD is doing slightly better in terms of backtracks but not enough to outperform domWDeg in terms of runtime.

For the Magic Square plot against time, there is a notable bend at about the 50% mark in most of the curves which can be explained by the fact that half of the instances only have 10% of their cells prefilled and present a bigger challenge. Interestingly, the simpler dom heuristics performs better than IBS and domWDeg, the latter being unable to solve half the instances in the allotted time. In contrast with the Nonogram Problem, here the full impact initialization is a very heavy procedure due to the high number of variable-value pairs to probe ($\approx n^4$ that is in our instances $9^4 = 6561$). It is also worth noting that on the Cost-Constrained Rostering Problem, maxSD solves seven out of the ten instances backtrack-free and is the only heuristic solving every instance. Similarly for the TTPPV Problem, almost 90% of the instances are solved backtrack-free by that heuristic. Moreover six instances happen to be infeasible and maxSD exhibits short proof trees for five of them, every other heuristic timing out on them.

### 7.11 Adding Randomized Restarts

It has been remarked that some combinatorial search has a strictly positive probability to reach a subtree that requires exponentially more time than the other subtrees encountered so far (so called "heavy-tail" behavior). Nonetheless, heavy tails can be largely avoided by adding randomized restarts on top of the search procedure (Gomes, Selman, & Kautz, 1998). This technique is orthogonal to the search heuristic employed and it systematically restarts the search every time a limit (typically a bound on the number of backtracks) is reached; obviously, in order to be effective, randomized restarts must be employed along with a heuristic that presents some sort of randomization or learning such that at each restart different parts of the search tree are explored. We tested the same heuristics to assess their performance with randomized restarts. The maxSD and IBS heuristics have been randomized: particularly, one variable-value pair is chosen at random with equal probability between the best two provided by the heuristic. Note that, as pointed out by Refalo (2004), impact information is carried over different runs to improve the quality of the impact approximation. As for domWDeg, the learned weights are kept between restarts. We implemented a slow geometric restart policy (Walsh, 1999) (that is $1, r, r^2, \ldots$ with $r = 2$) with a scale parameter optimized experimentally and separately for each problem type and search heuristic.

We turn again to Figure 7 and 8 but this time we also consider the curves for the heuristics with restarts. Restarts generally help a less informed heuristic such as dom, sometimes spectacularly so as for the Rostering Problem, but not always as indicated by the results on the Market Split Problem. For the other heuristics their usefulness is mixed: it makes little difference for maxSD except for Market Split where it degrades performance and Rostering where it improves its performance very significantly, now solving every instance very easily; for IBS it helps on the most difficult instances for half of the problems but for three others it degrades performance; for domWDeg it is generally more positive but never spectacular. Note that heavy-tail behavior of runtime distribution is conjectured to depend both on the problem structure and on the search heuristic employed (Hulubei & O'Sullivan, 2006). The Market Split Problem stands out as one where randomized restarts hurt every search heuristic considered.

### 7.12 Using Limited Discrepancy Search

Another way to avoid heavy tails is to change the order in which the search tree is traversed, undoing decisions made at the top of the search tree earlier in the traversal. A popular way of doing this is by applying limited discrepancy search (LDS) that visits branches in increasing order of their number of "discrepancies", which correspond to branching decisions going against the search heuristic (Harvey & Ginsberg, 1995). As for restarts, it can be combined with any search heuristic and may cause dramatic improvements in some cases but this less natural traversal comes with a price. Figure 9 illustrates the impact of LDS on two of our benchmark problems, using maxSD as the search heuristic. Either the usual depth-first search traversal is used ("maxSD" curve) or limited discrepancy search, grouping branches that have exactly the same number of discrepancies ("LDS 1"), by skips of 2 ("LDS 2"), or by skips of 4 ("LDS 4") discrepancies. On the rostering problem LDS undoes bad early decisions made by our heuristic and now allows us to solve every instance very quickly. However on the Magic Square problem the impact of LDS on the number of backtracks is low and it actually significantly slows down the resolution because LDS must revisit internal nodes, thus repeating propagation steps: the smaller the skip, the larger the computational penalty.

The same behavior could have been observed on other search heuristics and other problems. So LDS does not necessarily add robustness to our search.

### 7.13 Analyzing Variable and Value Selection Separately

One may wonder whether the success of counting-based search heuristics mostly depends on informed *value* selection, the accompanying variable selection being accessory. In order to investigate this, we introduce some hybrid heuristics:

- maxSD; random - selects a variable as in maxSD but then selects a value in its domain uniformly at random;
- IBS; maxSD - selects a variable as in IBS but then selects a value in its domain according to solution densities;
- domWDeg; maxSD - selects a variable as in domWDeg but then selects a value in its domain according to solution densities;

Figure 10 and 11 plot the number of solved instances against backtracks and time for our eight benchmark problems. Comparing maxSD and maxSD; random indicates that most of the time value selection according to solution densities is crucial, the Rostering Problem being an exception. Interestingly value selection by solution density improves the overall performance of IBS; for domWDeg it improves for the Latin Square and Magic Square problems but not for the rest, often decreasing performance. However such improvements do not really tip the balance in favor of other heuristics than maxSD, thus indicating that variable selection according to solution densities is also very important to its success.

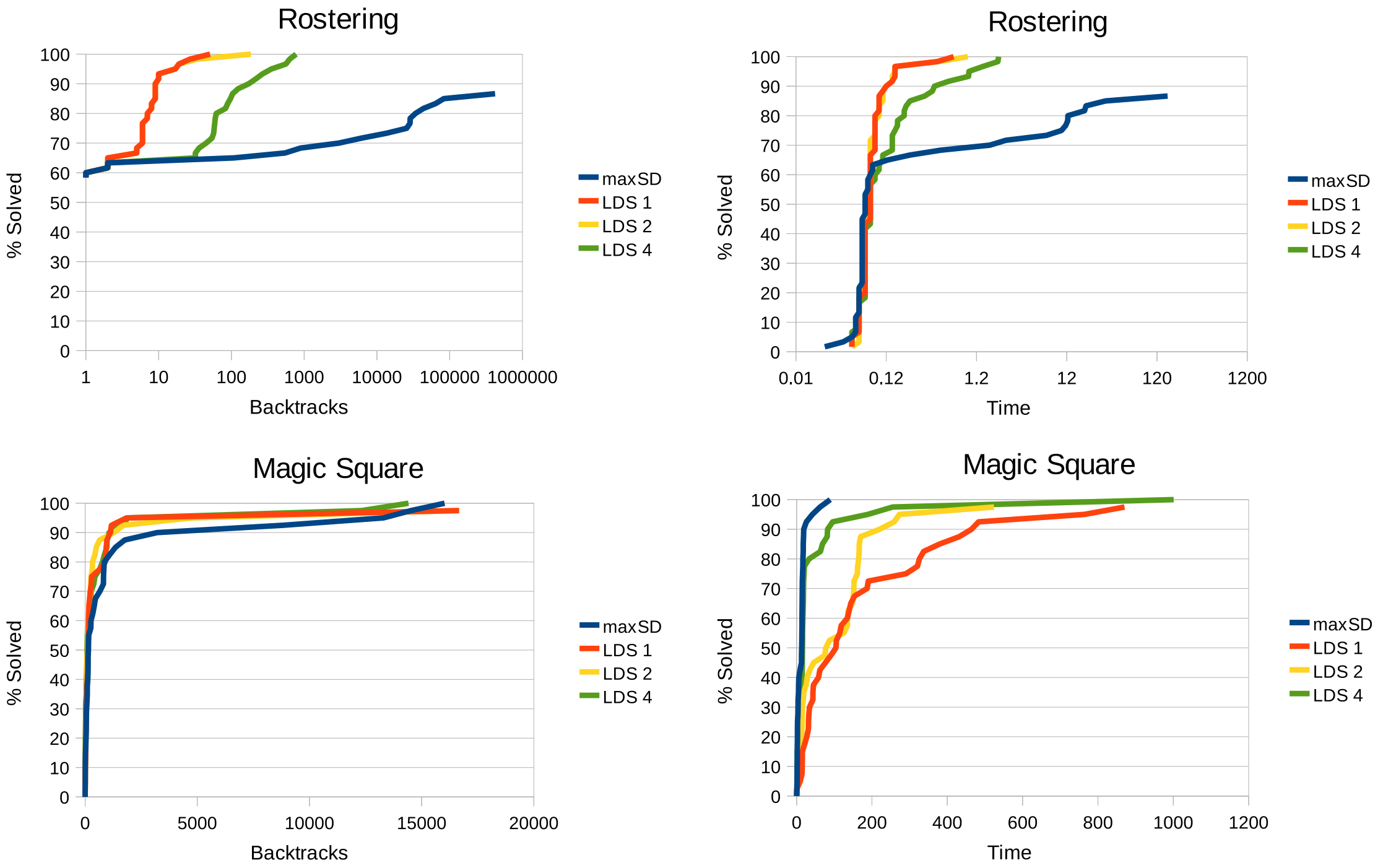

Figure 9: Percentage of solved instances with respect to the number of backtracks and to time (in seconds) for two of the benchmark problems. The maxSD search heuristic is used for every curve but the search tree traversal order is different.

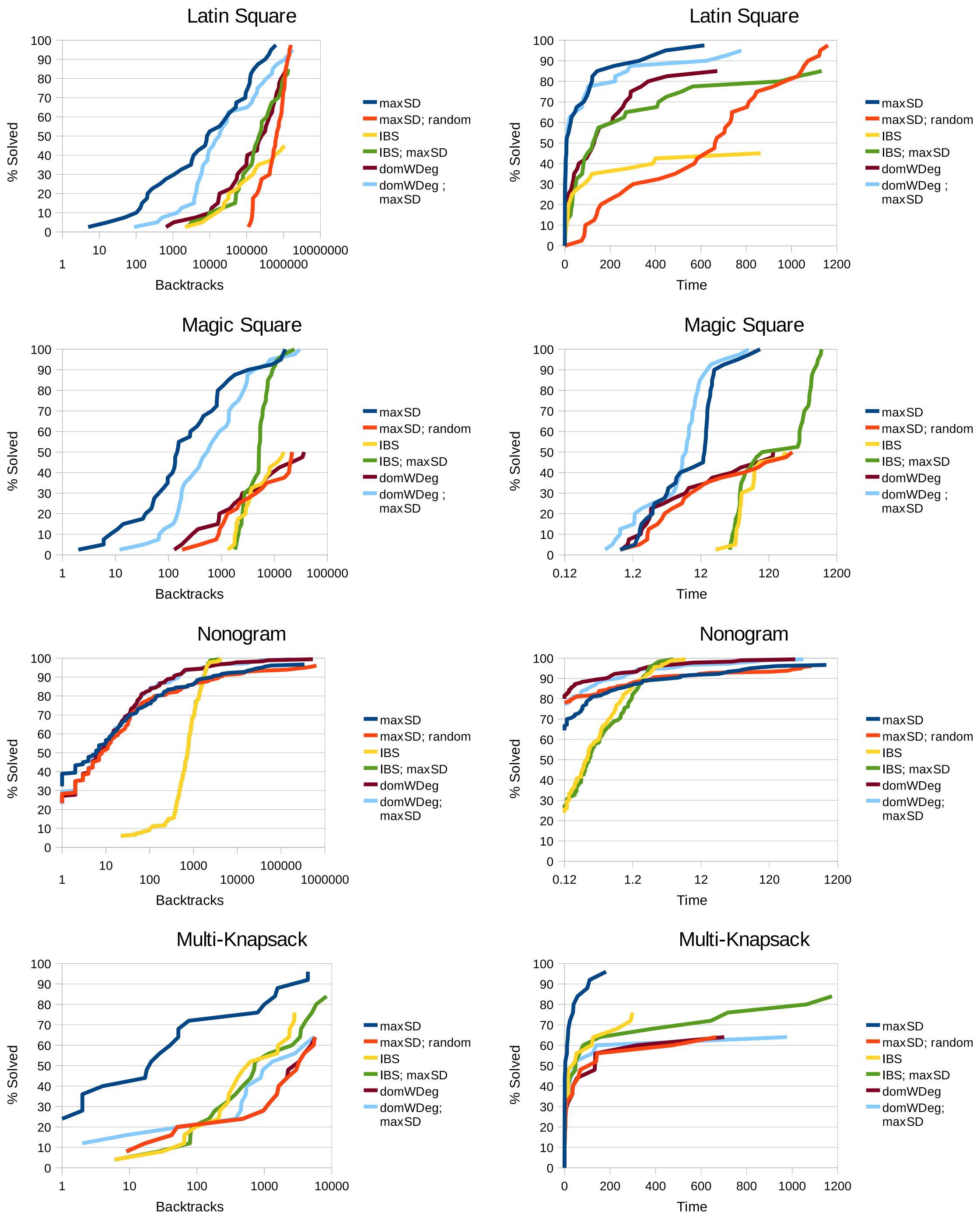

Figure 10: Percentage of solved instances with respect to the number of backtracks and to time (in seconds) for the first four benchmark problems. The search heuristics compared use solution densities either for variable or for value selection.

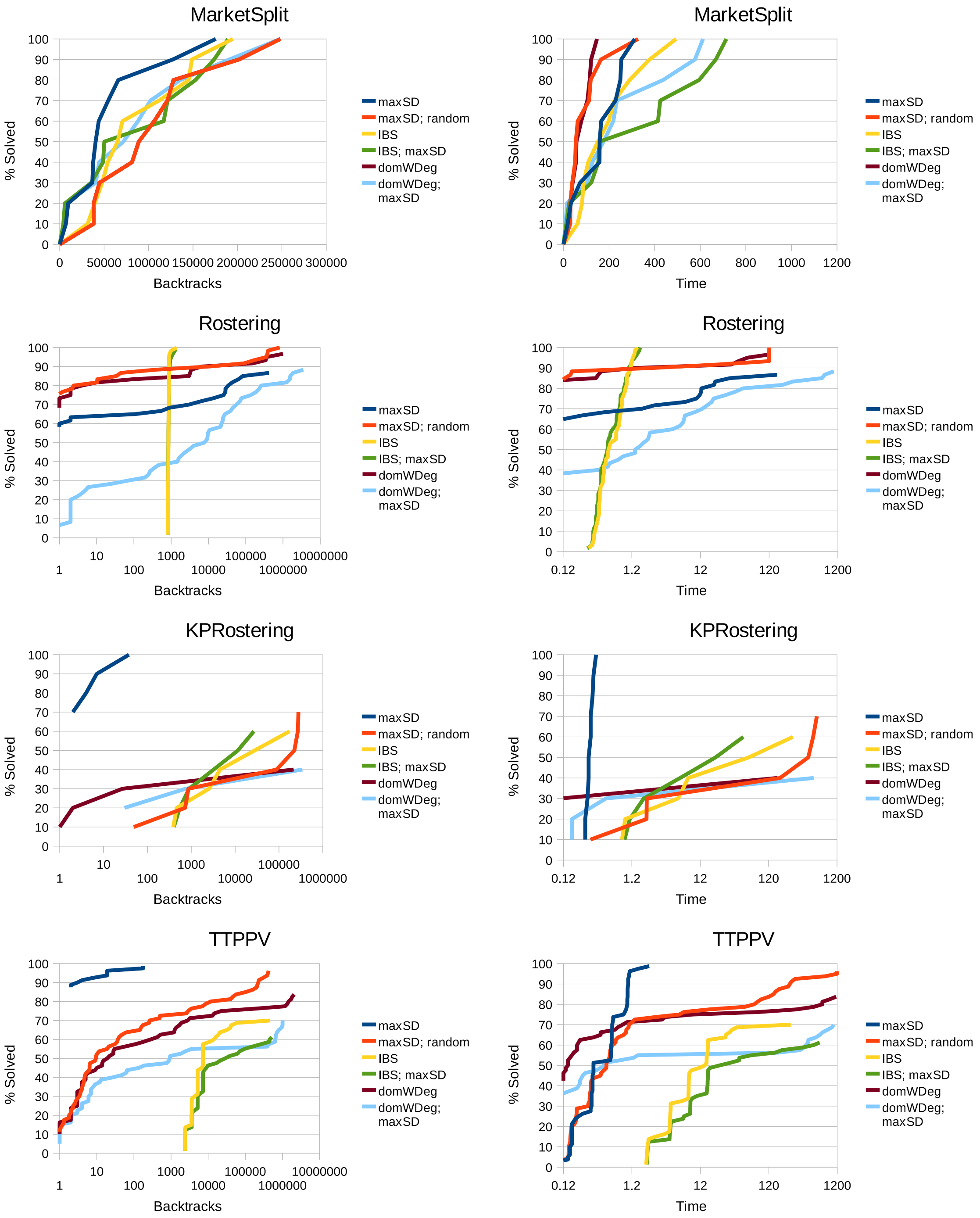

Figure 11: Percentage of solved instances with respect to the number of backtracks and to time (in seconds) for the last four benchmark problems. The search heuristics compared use solution densities either for variable or for value selection.

## 8. Conclusion

This paper described and evaluated counting-based search to solve constraint satisfaction problems. We presented some algorithms necessary to extract counting information from several of the main families of constraints in CP. We proposed a variety of heuristics based on that counting information and evaluated them. We then compared one outstanding representative, maxSD, to the state of the art on eight different problems from the literature and obtained very encouraging results. The next logical steps in this research include designing counting algorithms for some of the other common constraints and strengthening our empirical evaluation by considering new problems and comparing against application-specific heuristics. The next two paragraphs describe less obvious steps.

Users often need to introduce auxiliary variables or different views of the models that are linked together by channeling constraints. It is very important to provide all the counting information available at the level of the branching variables or at least at some level where direct comparison of solution densities is meaningful. For example in the case of the TTPPV an earlier model, in which two sets of variables each received solution densities from different constraints, did not perform nearly as well. Channeling constraints that express a one-to-many relation (such as the one present in the TTPPV) can be dealt with by considering value multiplicity in counting algorithms (Pesant & Zanarini, 2011). More complex channeling constraints represent however a limitation in the current framework.

Combinatorial *optimization* problems have not been discussed in this paper but are very important in operations research. Heuristics with a strong emphasis on feasibility (such as counting-based heuristics) might not be well suited for problems with a strong optimization component, yet may be very useful when dealing with optimization problems that involve hard combinatorics. Ideally, counting algorithms should not be blind to cost reasoning. One possibility that we started investigating not only counts the number of solutions that involve a particular variable-value pair but also returns the average cost of all the solutions featuring that particular variable-value pair. Another has shown promise when the cost is linear and decomposable on the decision variables (Pesant & Zanarini, 2011).

To conclude, we believe counting-based search brings us closer to robust automated search in CP and also offers efficient building blocks for application-specific heuristics.

## Acknowledgments

Financial support for this research was provided in part by the Natural Sciences and Engineering Research Council of Canada and the Fonds québécois de la recherche sur la nature et les technologies. We wish to thank Tyrel Russell who participated in the implementation and experimentation work. We also thank the anonymous referees for their constructive comments that allowed us to improve our paper.